\documentclass{article}

 \usepackage[eandd, final]{neurips_2026}

\usepackage[utf8]{inputenc} 
\usepackage[T1]{fontenc} 
\usepackage{wrapfig} 
\usepackage{hyperref}       
\usepackage{url}            
\usepackage{booktabs}       
\usepackage{amsfonts}       
\usepackage{nicefrac}       
\usepackage{microtype}      
\usepackage{xcolor}         
\usepackage{graphicx}
\usepackage{doi}

\usepackage{microtype}
\usepackage{graphicx}
\usepackage{subcaption}
\usepackage{booktabs} 
\usepackage{pifont}
\usepackage{multirow}
\usepackage{enumitem}
\usepackage{array}
\usepackage{graphicx} 

\usepackage{xcolor} 
\usepackage[table]{xcolor}
\usepackage{wrapfig}
\usepackage{needspace}

\newcommand{\cmark}{\ding{51}}
\newcommand{\xmark}{\ding{55}}

\usepackage{amsmath}
\usepackage{amssymb}
\usepackage{mathtools}
\usepackage{amsthm}
\usepackage{xurl}

\usepackage[capitalize,noabbrev]{cleveref}

\theoremstyle{plain}

\theoremstyle{definition}

\theoremstyle{remark}

\usepackage[textsize=tiny]{todonotes}
\title{~\raisebox{-8pt}{\includegraphics[height=30pt]{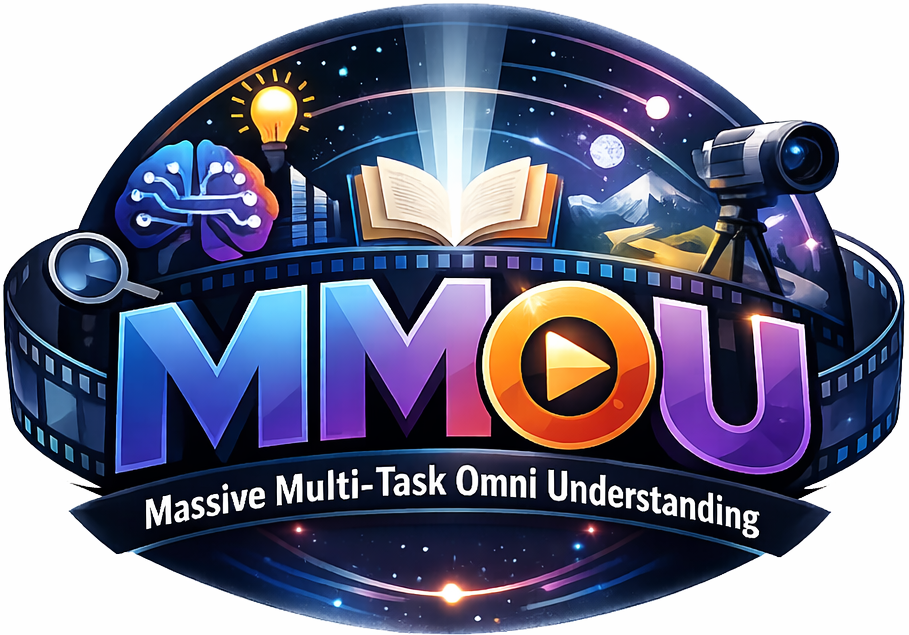}}~MMOU: A Massive Multi-Task Omni Understanding and Reasoning Benchmark for Long and Complex Real-World Videos}

\author{%
\begin{minipage}{0.98\textwidth}
\centering
\fontsize{9.6}{11.4}\selectfont
\newcommand{\asep}{\hspace{1.0em}}
\newcommand{\aname}[2]{\mbox{##1\textsuperscript{##2}}}
\aname{Arushi Goel}{1$\dagger$*}\asep
\aname{Sreyan Ghosh}{1,2$\dagger$*}\asep
\aname{Vatsal Agarwal}{2*}\asep
\aname{Nishit Anand}{2*}
\\[0.15em]
\aname{Kaousheik Jayakumar}{2$\ddagger$}\asep
\aname{Lasha Koroshinadze}{2$\ddagger$}\asep
\aname{Yao Xu}{1}\asep
\aname{Katie Lyons}{1}
\\[0.15em]
\aname{James Case}{1}\asep
\aname{Karan Sapra}{1}\asep
\aname{Kevin J. Shih}{1}\asep
\aname{Siddharth Gururani}{1}
\\[0.15em]
\aname{Abhinav Shrivastava}{2}\asep
\aname{Ramani Duraiswami}{2}\asep
\aname{Dinesh Manocha}{2}\asep
\aname{Andrew Tao}{1}
\\[0.15em]
\aname{Bryan Catanzaro}{1}\asep
\aname{Mohammad Shoeybi}{1}\asep
\aname{Wei Ping}{1}
\\[0.65em]
{\fontsize{9}{10.5}\selectfont
\textsuperscript{1} NVIDIA, USA
\qquad
\textsuperscript{2} University of Maryland, College Park, USA
}
\\[0.35em]
\end{minipage}%
}

\begin{document}
\maketitle
\begin{abstract}
\vspace{-2mm}
Multimodal Large Language Models (MLLMs) have shown strong performance in visual and audio understanding when evaluated in isolation. However, their ability to jointly reason over omni-modal (visual, audio, and textual) signals in long and complex videos remains largely unexplored. We introduce \textbf{MMOU}, a new benchmark designed to systematically evaluate multimodal understanding and reasoning under these challenging, real-world conditions. MMOU consists of 20,000 carefully curated questions paired with 11877 web-collected videos of varying length, spanning diverse domains and exhibiting rich, tightly coupled audio-visual content. The benchmark covers 13 fundamental skill categories, all of which require integrating evidence across modalities and time. All questions are manually annotated across multiple turns by professional annotators, ensuring high quality and reasoning fidelity. We evaluate 20+ state-of-the-art open-source and proprietary multimodal models on MMOU. The results expose substantial performance gaps: the best closed-source model achieves only 64.2\% accuracy, while the strongest open-source model reaches just 46.8\%. Our results highlight the challenges of long-form omni-modal understanding, revealing that current models frequently fail to apply even fundamental skills in long videos. Through detailed analysis, we further identify systematic failure modes and provide insights into where and why current models break. Project: \href{https://mmou.pages.dev/}{https://mmou.pages.dev/}

\end{abstract}

\vspace{-5mm}
\section{Introduction}
\label{sec:intro}
\vspace{-3pt}
\begin{wrapfigure}{r}{0.5\linewidth}
    \vspace{-4mm}
    \centering
    \includegraphics[width=\linewidth]{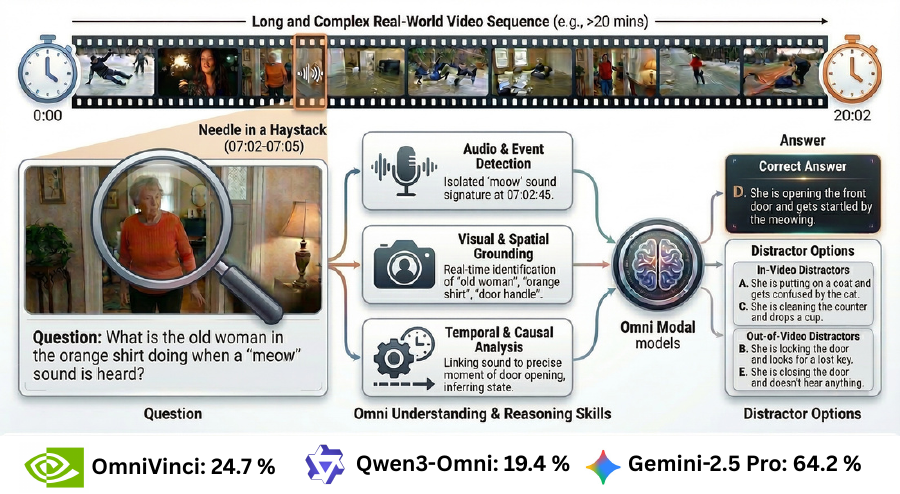}
    \caption{\small Overview of \textbf{MMOU}, a benchmark for evaluating omni-modal understanding in long, complex real-world videos, showing that both open and closed multimodal models struggle even with basic understanding.}
    \label{fig:radar_af2}
    \vspace{-4mm}
\end{wrapfigure}
The pursuit of Artificial General Intelligence (AGI) has driven rapid progress in Large Language Models (LLMs), particularly through the emergence of Multimodal Large Language Models (MLLMs) that process information across multiple modalities such as text, images, audio, and video~\citep{ye2025omnivincienhancingarchitecturedata, xu2025qwen3omnitechnicalreport, hurst2024gpt, comanici2025gemini, caffagni2024revolution}. These models have enabled compelling applications, allowing LLMs to see through vision~\citep{dai2024nvlm, liu2025nvilaefficientfrontiervisual, liu2023visualinstructiontuning} and listen through audio~\citep{goel2025audio, ghosh2025music, chu2024qwen2audiotechnicalreport, tang2024salmonngenerichearingabilities, tian2025ualmunifiedaudiolanguage}. Recent MLLMs demonstrate strong capabilities across audio tasks (e.g., automatic speech recognition, sound classification, and audio captioning) and visual tasks (e.g., OCR, visual question answering, and video grounding), often surpassing prior benchmarks by a large margin.

Despite this progress, existing MLLMs exhibit notable limitations. Most models are optimized for single-modality reasoning~\citep{bai2025qwen3vltechnicalreport, goel2025audio}, such as vision-only or audio-only understanding, and often fail to jointly perceive and reason across modalities analogous to human cognition. This limitation is partly due to the imbalance in available training data and benchmarks: single-modality datasets are more abundant, higher quality, and cover a wider range of tasks~\citep{liu2023llava, hurst2024gpt, google2023gemini} than their multi-modal counterparts. As a result, current models rarely learn to integrate audio and visual cues in a unified and consistent manner.

Benchmarking has long played a central role in advancing AI by providing structured, diagnostic evaluation frameworks~\citep{hendrycks2021measuringmassivemultitasklanguage, sakshi2024mmaumassivemultitaskaudio, kumar2025mmauprochallengingcomprehensivebenchmark, fu2025videommefirstevercomprehensiveevaluation, hu2025videommmuevaluatingknowledgeacquisition}. While evaluation of LLMs has matured substantially, covering domains such as mathematics, code generation, reasoning, and instruction following, holistic evaluation of MLLMs remains underdeveloped. Although numerous image and video benchmarks have emerged in recent years, benchmarks that rigorously evaluate audio-visual reasoning are scarce. In particular, most video benchmarks either ignore audio entirely or treat it as auxiliary, and predominantly focus on short clips that fail to capture long-term temporal dependencies~\citep{li2024mvbenchcomprehensivemultimodalvideo}. Consequently, existing evaluations do not adequately reflect the challenges posed by long and complex real-world videos, where meaningful understanding requires tightly coupled reasoning over audio and visual streams across extended time horizons.

\begin{figure*}[!t]
  \vskip 0.2in
  \begin{center}
    \centerline{\includegraphics[width=0.94\textwidth]{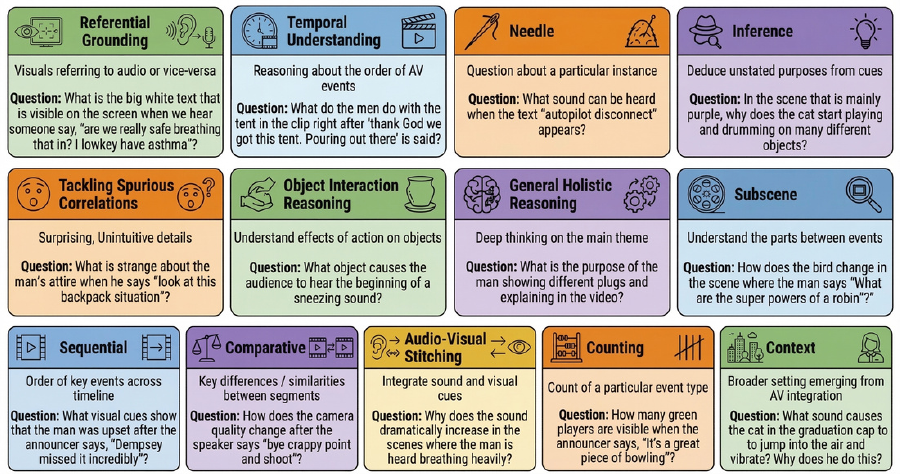}}
    \caption{\small Illustrative examples from \textbf{MMOU}, demonstrating the different skill types evaluated in the benchmark.}
    \label{fig:qa_example}
  \end{center}
  \vspace{-8mm}
\end{figure*}

{\noindent \textbf{Main Contributions.}} We present MMOU, a \textbf{M}assive, \textbf{M}ulti-task \textbf{O}mni-modal \textbf{U}nderstanding and Reasoning. Our benchmark is designed to evaluate joint audio-visual understanding and reasoning on long and complex real-world videos under realistic conditions (see Fig.~\ref{fig:radar_af2}). Specifically, (i) each question requires simultaneous integration of audio and visual information, such that removing either modality leads to failure; (ii) the questions require models to demonstrate proficiency in 13 distinct and fundamental skills (as shown in Fig.~\ref{fig:qa_example}); (iii) the benchmark is large-scale, comprising 20,000 multiple-choice QA pairs sourced from 11877 long-form real-world videos spanning 10 domains and 35 fine-grained subcategories, with each video exhibiting strong temporal and semantic alignment between audio and visual streams; and (iv) all questions are annotated by a group of 11 professionally trained human experts and each is optionally paired with 10 carefully constructed answer options that include hard distractors. To summarize, our main contributions are:

\begin{itemize}
\vspace{-3mm}
\setlength\parskip{0em}
    \item   We introduce \textsc{MMOU}, a comprehensive benchmark for evaluating advanced omni-modal (audio-visual) perception and reasoning in MLLMs on \emph{long and complex real-world videos}. MMOU spans 13 skill categories and includes 20,000 expertly annotated multiple-choice questions, covering both breadth and depth in multimodal understanding.

    \item We evaluate 20+ open-source and proprietary MLLMs on MMOU and show that even the most advanced models struggle with tasks that humans find intuitive. The best closed-source model achieves only 64.2\% accuracy, with open-source models performing substantially worse (46.8\%), revealing significant gaps in current multimodal reasoning capabilities.

    \item We conduct an in-depth analysis of model predictions, uncovering systematic failure modes.

\end{itemize}

\section{Related Work}
\label{sec:related_work}
\noindent
\textbf{Multimodal Large Language Models. }Recent years have seen rapid progress in multimodal large language models (MLLMs), which extend the capabilities of text-only LLMs~\citep{hurst2024gpt, meta2024llama3, yang2025qwen3} to visual, audio, and audio–visual inputs~\citep{xu2025qwen3omnitechnicalreport, goel2025audio,dai2024nvlm, bai2025qwen25vl, cheng2024videollama2,  xu2025qwen25omni}. These models typically integrate modality-specific encoders~\citep{xu2023metaclip, radford2021clip, ghosh2025audio, radford2023robust} with a shared language model backbone~\citep{chu2024qwen2, meta2024llama3, hurst2024gpt}, and are trained using large-scale multimodal instruction-tuning data~\citep{li2024llavaonevision, zhang2024llavavideo, goel2025audio, xu2025qwen3omnitechnicalreport}. As a result, state-of-the-art models demonstrate strong performance on a wide range of established benchmarks, including image–text, video–text, and audio–text understanding tasks~\citep{fu2024videomme, sakshi2024mmau, yue2024mmmu}.

Despite these advances, existing evaluation protocols remain largely unimodal, with most benchmarks isolating a single modality or task. Such narrowly defined settings fail to capture the complexity of real-world multimodal reasoning. Consequently, strong results on individual benchmarks do not necessarily translate to robust omni-modal understanding, which requires joint reasoning across modalities, tasks, and temporal context~\citep{li2024mvbench}. 


\noindent
\textbf{Multimodal Benchmarks}. A wide range of benchmarks have been proposed to evaluate multimodal models, including visual question answering~\citep{antol2015vqa}, video understanding~\citep{fu2024videomme, hu2025videommmuevaluatingknowledgeacquisition}, general image understanding~\citep{yue2024mmmu, masry2022chartqa, sidorov2020textcaps}, and audio reasoning~\citep{ma2025mmar, sakshi2024mmau, kumar2025mmauprochallengingcomprehensivebenchmark}. While these benchmarks have driven substantial progress, they predominantly evaluate isolated modalities or single-task settings, resulting in an incomplete evaluation of multimodal capabilities. Several audio-visual datasets such as VALOR~\citep{chen2023valor}, AVQA~\citep{yang2022avqa}, MusicAVQA~\citep{li2022learning}, AV-Odyssey~\citep{gong2024av}, AVHBench~\citep{sung2024avhbench}, AVCaps~\citep{sudarsanam2025avcaps} have been proposed for joint evaluation of multimodal models. More recent benchmarks such as WorldSense~\cite{hong2025worldsenseevaluatingrealworldomnimodal}, DailyOmni~\cite{zhou2025dailyomni}, OmniBench~\cite{li2024omnibench}, OmniVideoBench~\cite{li2025omnivideobenchaudiovisualunderstandingevaluation}, and UNO-Bench~\cite{chen2025unobenchunifiedbenchmarkexploring} move towards more complex joint audio–visual evaluation, but remain constrained in critical ways. They often limit questions to a single dominant modality~\citep{hong2025worldsenseevaluatingrealworldomnimodal, yang2022avqa, li2022learning, li2024omnibench}, focus on short-duration videos~\citep{zhou2025dailyomni, benchekroun2023worldsense}, or operate at a small scale with limited task diversity and category coverage~\citep{chen2025unobenchunifiedbenchmarkexploring, li2025omnivideobench}, preventing rigorous evaluation of long-context reasoning and joint cross-modal inference.

\vspace{-2.5mm}
\section{MMOU}
\label{sec:approach}

\begin{figure*}[!t]
  \begin{center}
    \centerline{\includegraphics[width=\textwidth]{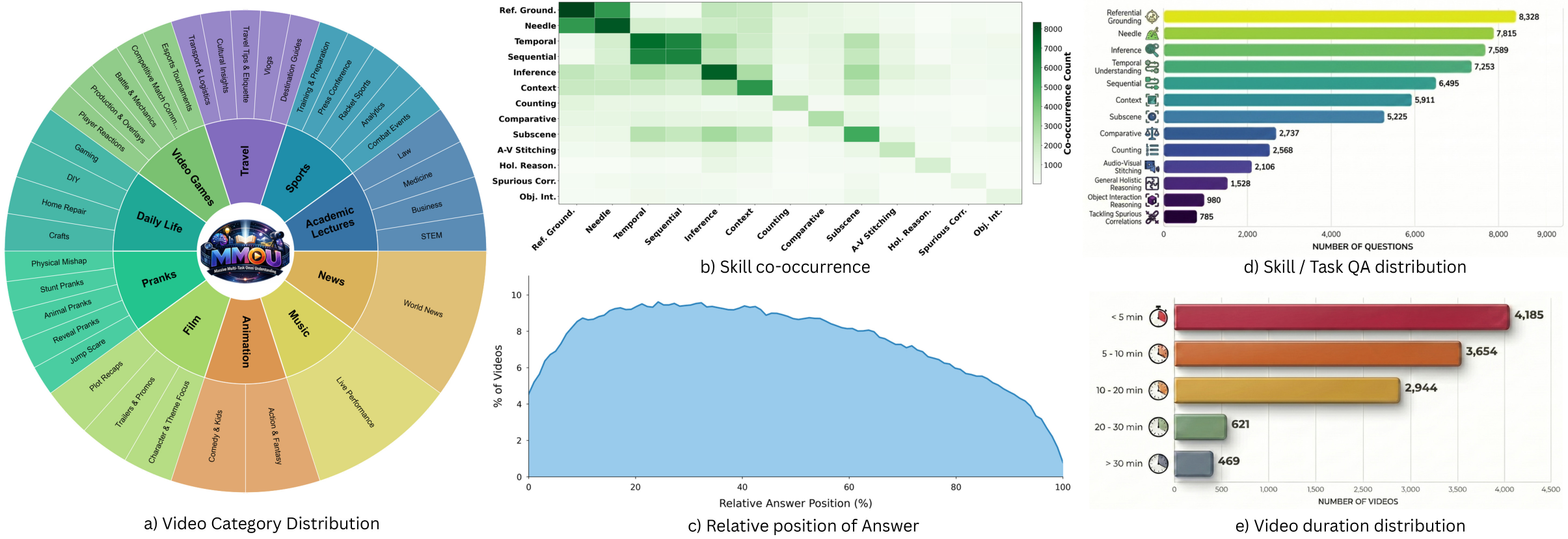}}
    \caption{\small Distribution of MMOU. (a) Video category distribution in MMOU, covering 10 major domains and 35 fine-grained subdomains. (b) Co-occurrence matrix of QA task types, illustrating how multiple reasoning skills are jointly required within individual questions. (c) Distribution of the relative temporal positions (average of start–end time-stamps) of answer evidence within videos, showing that answers are spread across the entire video timeline. (d) Distribution of QA instances across the 13 skill/task types in MMOU. (e) Video duration distribution, highlighting the prevalence of long and complex real-world videos.}
    \label{fig:qa_distribution}
  \end{center}
  \vspace{-10mm}
\end{figure*}
In this section, we first provide detailed statistics of MMOU in Section~\ref{subsec:data_stats} and compare it with previous benchmarks in Section~\ref{subsec:dataset_compare}. This is followed by a description of the data collection and annotation processes in Section~\ref{subsec:data_collection}.
\subsection{Dataset Statistics}
\label{subsec:data_stats}


\begin{wraptable}{r}{0.5\textwidth}
\vspace{-11mm}
\centering
\caption{Detailed statistics of MMOU.}
\label{tab:dataset_stats}
\resizebox{\linewidth}{!}{%
\begin{tabular}{l c | l c}
\toprule
\multicolumn{2}{c|}{\textbf{Video Statistics}} &
\multicolumn{2}{c}{\textbf{Annotation Statistics}} \\
\cmidrule(lr){1-2} \cmidrule(lr){3-4}
& \textbf{Test:Test-Mini} & & \textbf{Test:Test-Mini} \\
\midrule
\#Videos             & 9038:2841     & \#QAs                    & 15000:5000      \\
Major Categories     & 10:10         & Skill Types              & 13:11          \\
Subcategories        & 35:30         & Avg. Skills per QA       & 2.71:3.75      \\
Avg. Duration (s)    & 522.9:754.8   & Avg. Question Len. (w)   & 26.22:34.78    \\
Min. Duration (s)    & 7.0:13.0      & Avg. Answer Len. (w)     & 25.53:69.54    \\
Max. Duration (s)    & 7255:3586     & Avg. Answer Position (s) & 302.36:357.37  \\
\bottomrule
\end{tabular}}
\vspace{-4mm}
\end{wraptable}

Table~\ref{tab:dataset_stats} summarizes the key statistics of \textsc{MMOU}. The benchmark consists of 20,000 multiple-choice QA pairs, divided across test (15K) and test-mini (5K) subsets, collected from 11877 long-form real-world videos sourced from the web. Our videos are long, with an average duration of 522.9/754.8 seconds, a minimum of 7.0/13.0 seconds, and a maximum of 7255/3586 seconds for test/test-mini. All videos are sampled at 720p and hand-curated to promote content and domain diversity.

The videos span 10 major categories and 35 fine-grained subcategories, covering diverse domains such as academic lectures, sports, and other real-world scenarios (see Fig.~\ref{fig:qa_distribution}). Each question in MMOU is annotated with one or more of 13 skill types, with an average of 3 skills per question. A detailed breakdown of skill-wise question distribution is provided in Fig.~\ref{fig:qa_distribution}. All questions are initially annotated in an open-ended format. We subsequently convert them into a multiple-choice setting by constructing 9 hard distractors per question, resulting in 10 answer options per QA, as described in Section~\ref{subsec:data_collection}. The distribution of correct answer options is approximately uniform across all choices (\texttt{A}–\texttt{J}), as summarized in Table~\ref{tab:option_dist}, ~\ref{tab:option_dist_mini}. 

To avoid positional biases, where models may exploit answers appearing near the beginning or end of the video~\citep{liu2024lost,yuan2025cg}, we deliberately frame QAs with answer-relevant evidence at diverse temporal locations during annotation. As shown in Table~\ref{tab:dataset_stats}, the average answer position is 302.36 seconds, with its distribution relative to video length illustrated in Fig.~\ref{fig:qa_distribution}.

\subsection{Dataset Comparison}
\label{subsec:dataset_compare}

Table~\ref{tab:comparison} compares \textsc{MMOU} with existing multimodal benchmarks. Benchmarks such as AV-Odyssey and OmniBench primarily focus on single images paired with audio, whereas \textsc{MMOU} targets \emph{real-world videos with synchronized audio}, requiring joint audio-visual understanding. Compared to other omni-modal benchmarks, including DailyOmni, WorldSense, and OmniVideoBench, \textsc{MMOU} features substantially \emph{longer and more complex videos}, spanning durations from a few seconds to several hours, far exceeding the temporal scope of prior benchmarks.

To further validate the necessity of cross-modal reasoning, we randomly sample 20\% of MMOU and manually evaluate the instances. We find that this subset satisfies 100\% answer correctness and 100\% strict audio-visual dependency, substantially exceeding the cross-modal rigor of existing benchmarks reported in~\citet{chen2025unobenchunifiedbenchmarkexploring}. Additionally, we highlight that modality-specific models perform poorly on MMOU. As shown in Table~\ref{tab:domain_duration_eval}, the vision-only Qwen3-VL-32B achieves only 44\% accuracy, while the audio-only Qwen3-Omni attains 35.6\%, confirming that unimodal reasoning is insufficient. Overall, MMOU poses a significantly greater challenge than prior omni-modal benchmarks: even the widely used Qwen3-Omni-30B-A3B-Thinking model reaches only 19.4\% accuracy, markedly lower than its performance on existing benchmarks.

\begin{table*}[!t]
\centering
\caption{\small Comparison of MMOU with image (I), audio (A), video (V), and omni-modal QA benchmarks, highlighting MMOU’s scale, long-form videos and question type (Multiple Choice / Open Ended) in addition to strong audio-visual correspondence. $^*$ denotes that only the number of available videos is reported.}
\label{tab:comparson}
\small
\resizebox{\textwidth}{!}{%
\begin{tabular}{l c c c c c c c}
\toprule
\textbf{Benchmarks} & \textbf{Modality} & \textbf{\#Videos/Audios} & \textbf{Avg. Len.} &
\textbf{\#QA Pairs} & \textbf{\#Skills} & \textbf{QA Type} & \textbf{Open domain} \\
\midrule
MMMU     & I & N/A & N/A & 11500 & 32 & MC/Open  & \cmark \\
MMVU & V & 1529 & 51.4 & 3000 & 27 & MC/Open &  \xmark\\
MMAU       & A  & 9000 & 10.1 & 10000 & 27 & MC  & \cmark \\
MMAU-Pro      & A & 5787 & 123.8 & 5305 & 49 & MC/Open  & \cmark\\
Video-MME     & V  & 900 & 1017.9 & 2700 & 30 & MC  & \cmark \\
VideoMMMU & V & 300 & 506.2 & 900 & 30 & MC &  \xmark\\
LongVideoBench & V & 3763 & 473.0 & 6678 & 17 & MC &  \xmark\\
\midrule  

OmniBench     & A+I & 1142 & 9.17 & 1142 & 8 & MC &  \xmark \\
AV-Odyssey    & A+V+I & 620$^*$ & 15.58 & 4555 & 26 & MC &  \xmark \\

UNO-Bench & A+V+I & 384$^*$ & 27.1 & 1250 &  44 & MC/Open &  \cmark \\
DailyOmni & A+V & 684 & 43.7 & 1197  & 6 & MC &  \cmark \\

WorldSense & A+V & 1662 & 141.1 & 3172 & 26 & MC &  \cmark \\
OmniVideoBench & A+V &628 & 384.2 & 1000 & 13 & MC &  \cmark \\
\midrule
\textbf{MMOU (Ours)} & A+V  & 11877 & 578.4 & 20000 & 13 & MC/Open &  \cmark \\
\bottomrule
\end{tabular}%
\label{tab:comparison}
}
\end{table*}

\subsection{Data Collection, Curation \& Annotation}
\label{subsec:data_collection}

\begin{figure*}[!t]
  \vskip 0.2in
  \begin{center}
    \centerline{\includegraphics[width=0.96\textwidth]{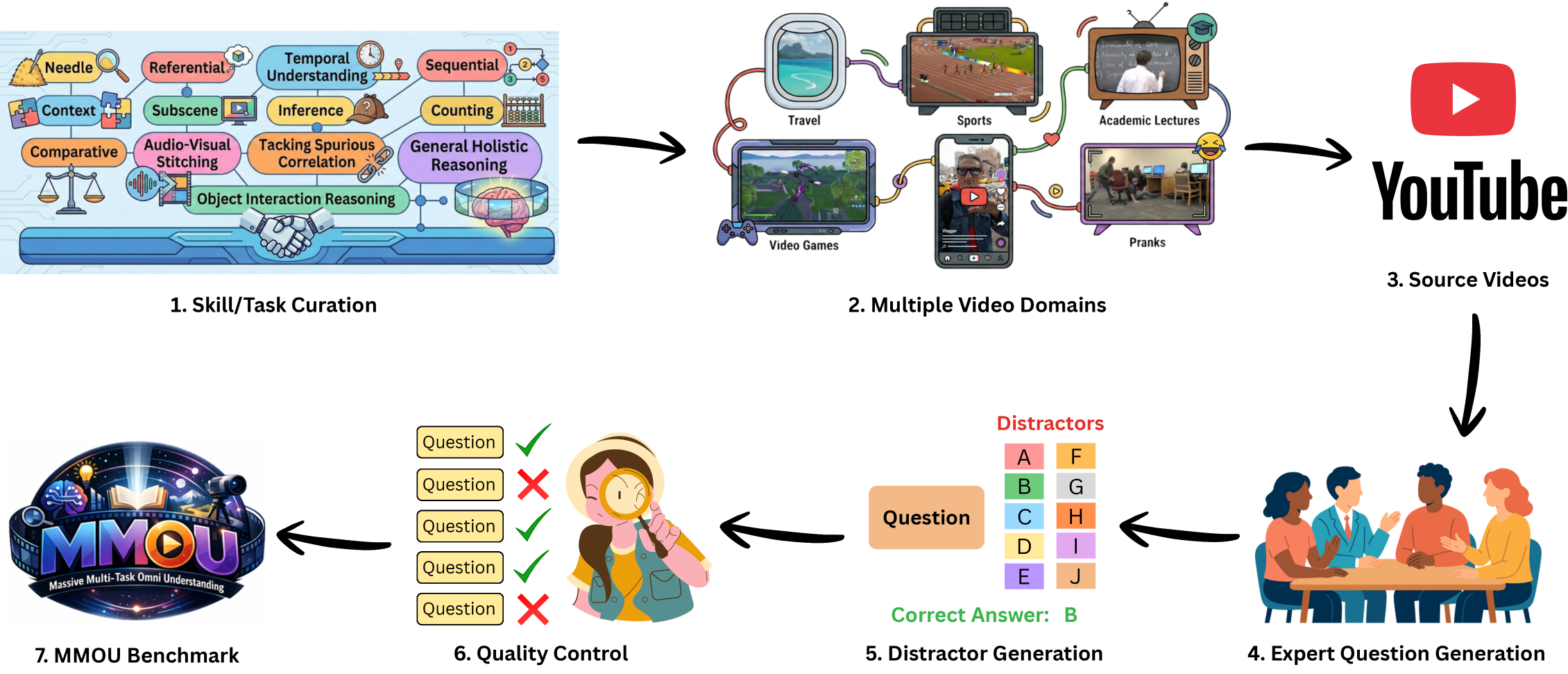}}
    \caption{
      Overview of the dataset-construction pipeline for MMOU.
    }
    \label{fig:dataset_creation}
  \end{center}
  \vspace{-0.4in}
\end{figure*}

Figure~\ref{fig:dataset_creation} illustrates the data construction pipeline for MMOU. We follow a structured, expert-driven process to ensure that all QAs require joint audio-visual understanding and reasoning over long, complex real-world videos.

\noindent \textbf{1. Skill and Task Curation.} First, we define a taxonomy of 13  fundamental audio-visual reasoning skills to capture the diverse challenges posed by long-form, real-world videos. These skills are designed to require explicit integration of audio and visual information and reflect the annotation ontology followed by expert annotators. 

\emph{Temporal understanding} and \emph{event sequencing} assess a model’s ability to reason about the order, progression, and temporal dependencies of audio-visual events across a video. \emph{Sub-scene understanding} focuses on identifying and interpreting semantically important segments within long videos, often requiring contextual understanding of surrounding events. \emph{Holistic video reasoning} evaluates global comprehension of the video’s main activity, objective, or theme, requiring integration of information across the entire timeline. \emph{Inference} and \emph{context understanding} require models to deduce unstated intentions, causes, or situational context from multiple audio-visual cues. \emph{Needle-in-the-haystack reasoning} tests the ability to localize and reason about specific moments in long videos, while \emph{referential grounding} evaluates linking between audio references and visual entities (or vice versa). \emph{Counting} and \emph{comparative reasoning} assess quantitative and relational reasoning over repeated or distinct audio-visual events. \emph{Object interaction reasoning} examines the understanding of actions performed on objects and their resulting transformations over time. \emph{Audio-visual stitching} evaluates reasoning over edited or stitched segments, requiring understanding of narrative continuity and editing intent. Finally, \emph{tracking spurious correlations} captures cases where correct answers rely on surprising or unintuitive audio-visual evidence that cannot be inferred from language priors alone. All questions are additionally tagged with \emph{audio-visual understanding}, ensuring that every instance requires joint reasoning over both modalities; questions solvable from a single modality are explicitly excluded. We provide examples in Table~\ref{tab:tasks_part1} and ~\ref{tab:tasks_part2}.
\vspace{-1mm}

\noindent \textbf{2. Video Domain Selection.} Guided by our curated skill taxonomy, we then systematically select a set of video domains to ensure broad coverage of real-world audio-visual understanding and reasoning scenarios. Specifically, we define 10 major video categories and 35 fine-grained subcategories, each chosen to exercise distinct combinations of the targeted skills. For each category and subcategory, we carefully curate videos to balance coverage across domains while maintaining sufficient diversity in content, temporal structure, and audio-visual dynamics. This domain-driven selection strategy ensures that MMOU spans a wide range of real-world contexts and supports comprehensive evaluation.
\vspace{-1mm}

\noindent \textbf{3. Source Video Collection.} We collect a total of 11877 real-world videos from publicly available online platforms (\textit{e.g.}, YouTube), with durations ranging from 7 seconds to 121 minutes. Videos are selected to align with the curated skill taxonomy, ensuring that each video supports the construction of at least one high-quality question. We prioritize naturally occurring content over scripted or synthetic data, resulting in realistic audio conditions, diverse visual scenes, and authentic temporal structure suitable for evaluating long-horizon audio-visual reasoning.
\vspace{-1mm}

\noindent \textbf{4. Expert Question Generation.} Eleven expert annotators follow a standardized annotation protocol. For each video, annotators first watch the video in its entirety. They then generate open-ended question–answer pairs that require \emph{joint} audio and visual understanding, explicitly avoiding yes/no questions or questions answerable from text alone. More detailed guidelines are present in Appendix~\ref{sec.ann_instructions}. Annotators are required to annotate the earliest and latest timestamps at which the supporting evidence for the answer appears, and are encouraged to diversify the same. Each question is tagged with one or more skill categories from our predefined taxonomy. We encourage annotators to generate multiple diverse questions per video, which are then filtered.

\noindent \textbf{5. Distractor Generation.} All questions are initially authored in an open-ended format. We then convert them into a multiple-choice setting by generating nine hard distractors per question, resulting in ten answer options. Distractors are generated using GPT-5.2~\cite{openai2025gpt52systemcard}, conditioned on the question and additional video-level metadata; the full prompt is provided in Fig.~\ref{fig:prompt_option_gen}. To increase difficulty, half of the distractors are designed to be semantically plausible and grounded in the video context, while the remaining half are intentionally out-of-context. This balanced construction prevents elimination via superficial cues and encourages genuine audio-visual reasoning. To further increase question difficulty and following prior work~\citep{tam2025none}, we replace the correct answer with \emph{``None of the above''} in 13\% (2000) of the test QAs, and one of the incorrect options in a further 13\% (2000). 
For test-mini, 250 QAs have the correct answer replaced with \emph{``None of the above''}, and 250 QAs have an incorrect option replaced with \emph{``None of the above''}.


\vspace{-1mm}

\noindent \textbf{6. Quality Control and Filtering.} A separate group of expert reviewers conducts rigorous quality control, removing ambiguous, redundant, or overly trivial questions, as well as instances with misaligned timestamps or weak audio-visual grounding. Only questions that strictly require joint audio-visual reasoning and adhere to the annotation guidelines are retained, resulting in a final set of 20,000 QA pairs. The resulting annotations show strong consistency, achieving a Fleiss' kappa score of 0.86. We also provide an empirical analysis of distractor quality in Appendix~\ref{sec:distractor_quality}.
\vspace{-1mm}

\noindent \textbf{7. MMOU Finalization.} The final \textsc{MMOU} benchmark consists of 20,000 carefully curated and reviewed QA instances.

\vspace{-3mm}
\section{Experimental Setup}
\label{sec:experiments}
\vspace{-2mm}
\begin{table*}[!t]
\centering
\caption{Performance breakdown across video domains and video durations for closed-source, open access, and open-source audio-visual MLLMs, video-only, audio-only MLLMs, and text-only LLMs.}
\label{tab:domain_duration_eval}
\small
\renewcommand{\arraystretch}{1.2}

\resizebox{\textwidth}{!}{
\begin{tabular}{l | c c c c c c c c c c | c c c c c | >{\columncolor{gray!15}\color{blue!60!black}}c}
\toprule 
&  
\multicolumn{10}{c|}{\textbf{Video Domains}} &
\multicolumn{5}{c|}{\textbf{Video Durations (min)}} &
\multicolumn{1}{c}{\cellcolor{gray!15}\textbf{\color{blue!60!black}Overall}} \\

\cmidrule(lr){2-11}
\cmidrule(lr){12-16}
\cmidrule(lr){17-17}

\textbf{Methods} &
{Sports} & {Travel} & {Video Games} & {Daily Life} & {Academic} &
{Film} & {Pranks} & {Music} & {Animation} & {News} &
{$<5$} & {5--10} & {10--20} & {20--30} & {$>30$} &
{Any} \\

\midrule
Random   & 10.0 & 10.0 & 10.0 & 10.0 & 10.0 & 10.0 & 10.0 & 10.0 & 10.0 & 10.0 & 10.0 & 10.0 & 10.0 & 10.0 & 10.0 & 10.0 \\
Human   & 86.3 & 85.7 & 82.7 & 85.1 & 83.5 & 85.0 & 83.9 & 86.1 & 82.0 & 90.0 & 87.2 & 85.6 & 84.0 & 83.0 & 81.5 & 84.3 \\
\midrule
\multicolumn{17}{c}{\textit{Closed-Source Audio-Visual MLLMs}} \\
\midrule
Gemini 2.5 Pro  & \textbf{61.2} & \textbf{67.3} & \textbf{60.9} & \textbf{68.1} & \textbf{71.4} & \textbf{66.5} & \textbf{71.0} & \textbf{59.7} & \textbf{58.2} & \textbf{61.8} & \textbf{62.2} & \textbf{66.2} & \textbf{66.2} & \textbf{59.0} & \textbf{58.5} & \textbf{64.2} \\
Gemini 2.5 Flash  & \underline{56.2} & \underline{59.1} & \underline{46.1} & \underline{60.2} & \underline{57.5} & \underline{61.1} & \underline{54.3} & \underline{52.1} & \underline{49.5} & \underline{52.9} & \underline{55.9} & \underline{57.4} & \underline{57.6} & \underline{49.8} & \underline{45.6} & \underline{55.8} \\
\midrule
\multicolumn{17}{c}{\textit{Open-Source Audio-Visual MLLMs}} \\
\midrule
Qwen2.5-Omni-7B  & 35.8 & 29.0 & 18.5 & 36.0 & 26.4 & 26.2 & 20.4 & 28.3 & 20.5 & 30.0 & 35.4 & 32.6 & 29.9 & 25.6 & 22.6 & 31.3 \\
Qwen3-Omni-30B-A3B-Instruct  & \underline{50.3} & \textbf{39.5} & \underline{28.3} & \textbf{51.6} & \underline{40.3} & \textbf{39.8} & \underline{27.4} & \underline{41.3} & \textbf{37.0} & \textbf{47.1} & \textbf{48.2} & \underline{47.9} & \textbf{44.9} & \textbf{38.0} & \textbf{43.6} & \underline{46.0} \\
Qwen3-Omni-30B-A3B-Thinking  & 20.3 & 19.8 & 14.6 & 20.3 & 23.9 & 22.8 & 11.8 & 13.8 & 21.2 & 18.9 & 20.4 & 20.1 & 18.9 & 16.5 & 18.2 & 19.4 \\
Phi-4 Multimodal  & 34.9 & 28.9 & 23.2 & 33.3 & 27.1 & 24.3 & 24.5 & 33.2 & 23.6 & 31.4 & 33.6 & 32.0 & 30.2 & 29.5 & \underline{27.9} & 31.4 \\
Gemma 3n & 36.6 & 23.5 & 19.4 & 35.7 & 23.4 & 24.9 & 26.3 & 29.0 & 24.6 & 28.6 & 33.8 & 31.3 & 29.3 & 27.0 & 27.5 & 30.7 \\
Minicpm-o 4.5 & \textbf{50.7} & \underline{39.3} & \textbf{30.4} & \underline{50.8} & \textbf{43.6} & \underline{36.0} & \textbf{35.1} & \textbf{43.3} & \underline{29.2} & \underline{46.3} & \underline{48.1} & \textbf{49.8} & \underline{39.2} & \underline{33.3} & 9.1 & \textbf{46.8} \\
Video-LLaMA 2 & 27.1 & 24.4 & 18.5 & 27.7 & 24.7 & 22.1 & 23.3 & 25.1 & 22.5 & 22.5 & 26.7 & 25.9 & 23.4 & 22.8 & 22.7 & 24.8 \\
OmniVinci  & 27.9 & 26.1 & 16.1 & 26.6 & 27.6 & 24.2 & 19.1 & 23.4 & 6.3 & 24.7 & 28.4 & 26.1 & 24.7 & 21.7 & 9.9 & 24.7 \\
Baichuan-Omni-1.5 & 27.9 & 24.5 & 19.5 & 25.4 & 21.9 & 19.9 & 16.9 & 23.8 & 17.2 & 23.3 & 28.9 & 25.2 & 20.0 & 14.6 & 8.5 & 23.2 \\

\midrule
\multicolumn{17}{c}{\textit{Video-Only Multimodal MLLMs}} \\
\midrule
Qwen3-VL-32B-Instruct & \textbf{47.8} & \textbf{39.9} & \textbf{31.9} & \textbf{48.4} & \textbf{37.2} & \textbf{40.4} & \textbf{41.9} & \textbf{45.5} & \textbf{44.1} & \textbf{42.2} & \textbf{44.5} & \textbf{45.3} & \textbf{43.3} & \textbf{40.4} & \textbf{44.1} & \textbf{44.0} \\
Qwen3-VL-8B-Instruct  & \underline{38.9} & \underline{33.3} & \underline{26.3} & \underline{40.6} & \underline{31.5} & \underline{34.9} & \underline{32.2} & \underline{36.4} & \underline{39.7} & \underline{33.8} & \underline{36.4} & \underline{36.8} & \underline{36.0} & \underline{33.1} & \underline{35.6} & \underline{36.1} \\
Qwen2.5-VL-7B-Instruct  & 34.2 & 28.1 & 21.5 & 34.9 & 24.8 & 23.7 & 24.6 & 31.0 & 27.4 & 27.6 & 32.0 & 31.4 & 29.0 & 26.2 & 27.6 & 30.2 \\
\midrule
\multicolumn{17}{c}{\textit{Audio-Only Multimodal MLLMs}} \\
\midrule
Audio Flamingo 3 
& \underline{18.7} & \underline{15.4} & \underline{13.0} & \underline{19.4} & \underline{15.6} & \underline{13.7} & \underline{11.1} & \underline{17.8} & \underline{12.3} & \underline{18.8}
& \underline{18.7} & \underline{19.1} & \underline{16.9} & \underline{16.2} & \underline{13.9} & \underline{17.7} \\
Qwen3-Omni-30B-A3B   & \textbf{35.9} & \textbf{37.3} & \textbf{28.0} & \textbf{36.4} & \textbf{38.9} & \textbf{36.7} & \textbf{33.1} & \textbf{44.4} & \textbf{36.4} & \textbf{50.0} & \textbf{40.6} & \textbf{36.7} & \textbf{33.7} & \textbf{35.1} & \textbf{34.5} & \textbf{35.6} \\

\midrule
\multicolumn{17}{c}{\textit{Cascaded Models}} \\
\midrule
Qwen3-(VL+O-A) + Qwen3-235B  & \textbf{34.5} & \textbf{37.3} & \textbf{26.4} & \textbf{37.3} & \textbf{39.0} & \textbf{31.4} & \textbf{20.8} & \textbf{24.7} & \underline{27.2} & \textbf{30.9} & \textbf{32.8} & \textbf{33.4} & \textbf{33.7} & \textbf{31.9} & \textbf{30.8} & \textbf{33.1} \\
Qwen3-(VL+O-A) + GPT-5.2  & \underline{28.8} & \underline{29.0} & \underline{21.6} & \underline{30.3} & \underline{36.8} & \underline{30.8} & \underline{20.7} & \underline{21.1} & \textbf{27.7} & \underline{26.5} & \underline{27.7} & \underline{28.6} & \underline{29.4} & \underline{25.0} & \underline{25.4} & \underline{28.1} \\
\midrule
\multicolumn{17}{c}{\textit{Text-Only LLMs}} \\
\midrule
Qwen3-235B  & \textbf{40.8} & \textbf{32.5} & \textbf{24.2} & \textbf{38.7} & \textbf{29.7} & \textbf{34.4} & \textbf{28.4} & \textbf{39.2} & \textbf{34.2} & \textbf{39.6} & \underline{40.5} & \textbf{37.6} & \textbf{36.4} & \textbf{32.8} & \textbf{36.4} & \textbf{37.5} \\
GPT-4o mini  & 22.9 & 28.3 & 16.8 & 22.2 & 26.8 & 23.6 & 22.2 & 25.5 & 28.1 & 20.0 & 26.3 & 23.7 & 23.6 & 22.0 & 23.2 & 23.8 \\
GPT-4.1  & 37.4 & 30.7 & 19.6 & 38.3 & 28.6 & 27.7 & 22.2 & 34.0 & 27.2 & 34.0 & 35.5 & 35.0 & 33.1 & 28.8 & 33.3 & 33.9 \\
\bottomrule
\end{tabular}
}
\vspace{-3mm}
\end{table*}

\subsection{Baselines}
\label{subsec:baselines}
\vspace{-3mm}

We evaluate MMOU on a diverse set of baselines spanning omni-modal, audio-only, vision-only, and text-only models.

\noindent \textbf{Audio-Visual Multimodal Large Language Models.} We evaluate SOTA large omni-modal models that are explicitly designed to jointly process audio and visual inputs. These models integrate modality-specific encoders with a shared language backbone and are trained using large-scale multimodal instruction-tuning data. We include both closed-source and open-source omnimodal models. Specifically, the closed-source baselines include Gemini~2.5~Flash and Pro \citep{comanici2025gemini}. The open-source omni-modal models evaluated are Qwen~2.5-Omni \citep{xu2025qwen25omni}, Qwen~3-Omni-Instruct, Qwen~3-Omni-Think \citep{xu2025qwen3omnitechnicalreport}, Phi-4~Multimodal \citep{abouelenin2025phi}, Gemma~3n \citep{gemmateam2025gemma3technicalreport}, MiniCPM \citep{openbmb2025minicpm-o2.6}, Video-LLaMA~2 \citep{cheng2024videollama2}, OmniVinci \citep{ye2025omnivincienhancingarchitecturedata}, and Baichuan-Omni \citep{li2025baichuan}.

\noindent \textbf{Audio-only and Vision-Only MLLMs.} To isolate the contributions of visual and audio cues, we additionally evaluate MMOU using modality-restricted models. For vision-only large vision–language models, we consider Qwen3-VL-32B-Instruct and Qwen3-VL-8B-Instruct~\citep{bai2025qwen3vltechnicalreport} and Qwen2.5-VL-7B-Instruct~\citep{bai2025qwen25vl}. For audio-only evaluation, we include Audio Flamingo 3~\citep{goel2025audio} and Qwen3-Omni-Instruct~\citep{xu2025qwen3omnitechnicalreport} operating in audio-only mode. 


\noindent \textbf{Text-Only Large Language Models \& Cascaded Models.}
Finally, we evaluate text-only large language models and text-centric reasoning baselines. We employ Qwen3-235B, GPT variants, and only pass the question and options without any audio or visual inputs. In addition, we consider two cascaded caption-based baselines. For this setup, we first generate audio and visual captions of the video separately using Qwen3-Omni-30B-A3B and Qwen3-VL-235B-A22B-Instruct, respectively. The generated captions are then fused into a single coherent audio-visual description of the video, which is then provided to a text-only LLM to answer the question. This design evaluates whether text descriptions alone are sufficient for solving MMOU in the absence of multimodal perception.

\noindent
\textbf{Evaluation}
We evaluate our models using micro-averaged accuracy. For each question, models are shown a set of answer options and instructed to select exactly one. Next, we apply robust regular-expression–based parsing to extract the predicted option and match it via string comparison. To reduce option-order bias, we randomize the option order five times and take the majority-selected answer. We further evaluate each model under multiple prompt variants and report the best-performing prompt configuration for all MLLMs.

\vspace{-3mm}
\section{Results and Discussion}
\vspace{-3mm}

In \Cref{tab:domain_duration_eval}, we present results on the MMOU test-subset on 20+ open- and closed-source audio-visual MLLMs, LVLMs, LALMs, and text-only LLMs. In \Cref{tab:testmini_overall}, we present results on test-mini-subset. Proprietary closed-source Gemini 2.5 Pro~\citep{google2023gemini} establishes itself as the strongest baseline with an overall accuracy of 64.2\% across diverse video domains (sports, news, travel, etc.) and durations (short, medium, and long). Compared to the performance of other open-source audio-visual multimodal models (e.g., Qwen3Omni and OmniVinci), which experience a relative drop in performance of more than 24.7\%, we hypothesize the relatively strong performance of Gemini to pre-training on YouTube videos. Even the state-of-the-art models fall well short of human-level performance of 84.3\% posing fundamental challenges to joint audio-visual perception and reasoning.

\Needspace{18\baselineskip}
\begin{wraptable}[20]{r}{0.38\textwidth}

\centering
\caption{Performance of models on test-mini subset (overall accuracy).}
\label{tab:testmini_overall}
\small
\renewcommand{\arraystretch}{1.0}
\begin{tabular}{l | >{\columncolor{gray!15}\color{blue!60!black}}c}
\toprule
\textbf{Methods} & \textbf{\color{blue!60!black}Overall} \\
\noalign{\vskip -2pt}\midrule
Human & 82.60 \\
\noalign{\vskip -2pt}\midrule
Gemini 2.5 Pro & \textbf{78.58} \\
Gemini 2.5 Flash & \underline{66.21} \\
Qwen3-Omni-30B-Inst. & 54.10 \\
Qwen3-Omni-30B-Think & 35.76 \\
Phi-4 Multimodal & 36.40 \\
MiniCPM-V-4.6 & 32.90 \\
Gemma-3n-E4B-it & 32.06 \\
VideoLLaMA2 & 28.40 \\

OmniVinci & 27.75 \\

Qwen3-VL-32B-Inst. & 52.19 \\
Qwen3-VL-8B-Inst. & 41.90 \\
Qwen2.5-VL-7B-Inst. & 34.58 \\
Audio Flamingo 3 & 16.81 \\
Qwen3-Omni-30B-Audio & 36.06 \\
GPT4-mini & 41.64 \\
Qwen3-235B (Cascaded) & 47.91 \\
\bottomrule
\end{tabular}
\vspace{-12mm}
\end{wraptable}

\textbf{Cross-modal understanding is critical in MMOU.} To evaluate the importance of cross-modal reasoning, we benchmark several video-only baselines from the Qwen-VL series. Despite being the state-of-the-art model in complex vision tasks, Qwen3-VL-32B achieves a low performance of 44\%, necessitating the need for strong audio-visual integration. Similarly, state-of-the-art audio-only language models fail to answer most of the questions with audio modality alone, seeing a significant drop in performance of 17.7\% with Audio Flamingo 3 and 35.6\% with Qwen3‑Omni‑30B‑A3B (audio‑only). This confirms that MMOU requires both audio and visual modalities to answer the questions. \textbf{Additionally, we also perform several unimodal ablations on omni models in Section~\ref{app:unimodal_ablation} and show that inference on a single modality almost always underperforms inference on both, further strengthening our argument.}

\noindent \textbf{Text-only Large Language Models \& Cascaded Models.} Furthermore, we present an evaluation on SOTA text-only LLMs, Qwen3-235B~\citep{yang2025qwen3} and GPT-4~\citep{openai2025gpt52systemcard}, confirming minimal textual biases in the question and answer choices. Without audio-visual inputs, we cannot achieve state-of-the-art performance using commonsense knowledge and language biases alone. This effectively validates the dataset's design and the need for true, temporally grounded audio-visual perception and reasoning. Moreover, we benchmark cascaded models by fusing video and audio captions with the question as context to the LLM. Providing a rich contextual audio-visual summary is not sufficient and indicates the need for joint end-to-end cross-modal perception. 

\vspace{-2mm}
\section{Results Analysis}
\label{sec:results_analysis}
\vspace{-2mm}

\begin{figure}[!h]
    \centering
    \begin{subfigure}[b]{0.48\textwidth}
        \centering
        \includegraphics[trim=0.4cm 0.5cm 0cm 1.5cm, width=0.65\textwidth]{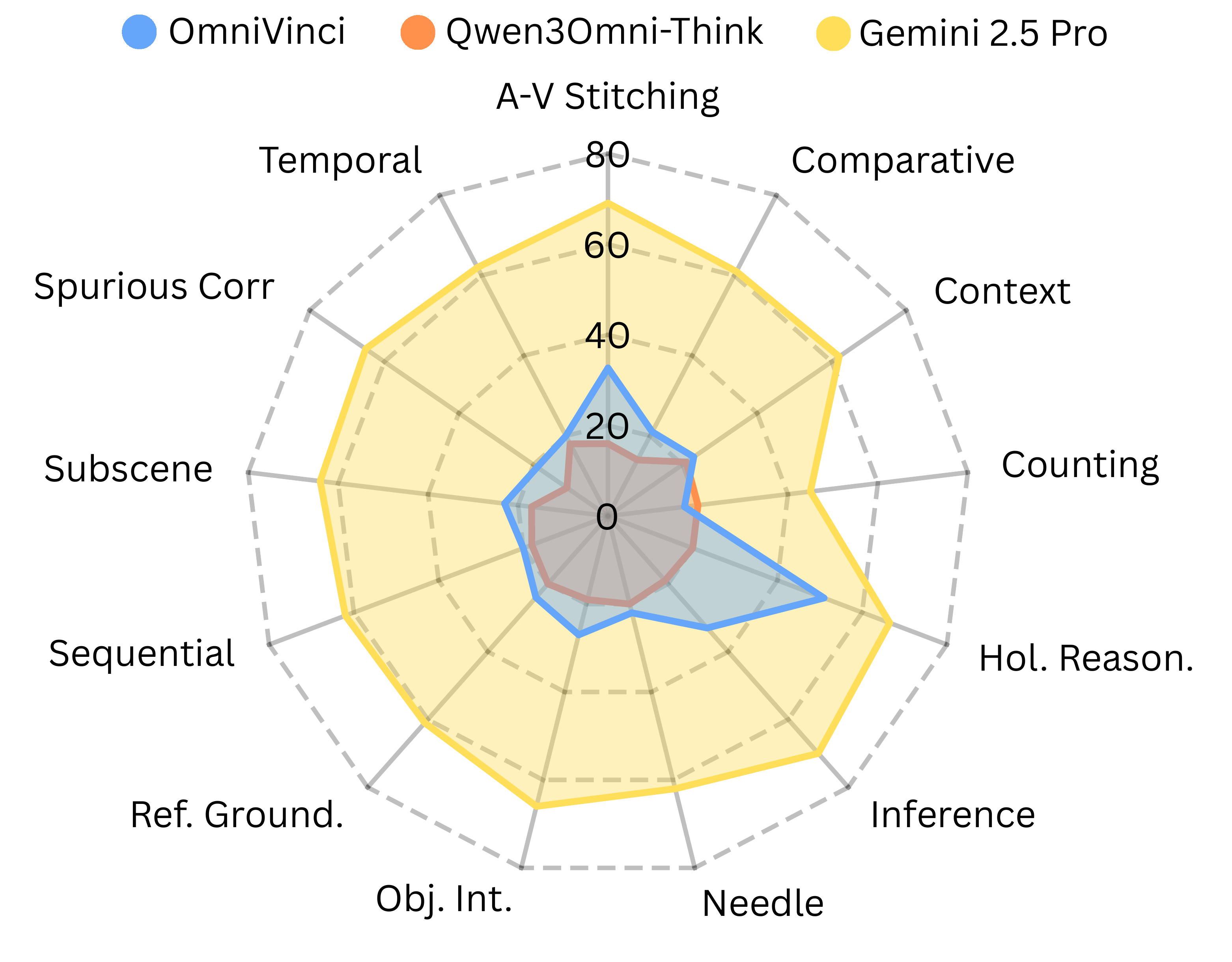}
        \caption{\small Skill-wise performance comparison of various models on MMOU Test set.}
        \label{fig:radar_results}
    \end{subfigure}
    \hfill
    \begin{subfigure}[b]{0.48\textwidth}
        \centering
        \includegraphics[width=\textwidth, trim=1cm 0.25cm 0cm 0cm]{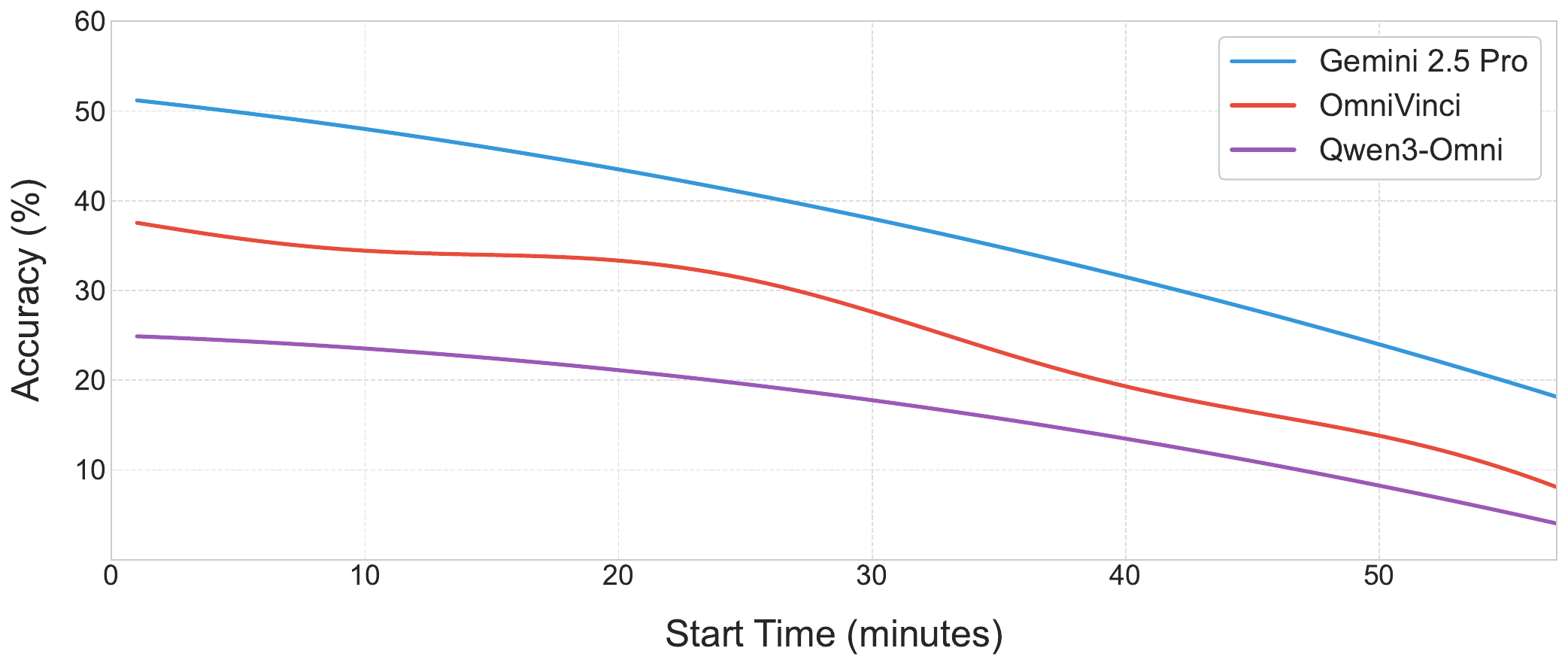}
        \caption{\small Model accuracy vs.\ evidence position in long videos.}
        \label{fig:time_line}
    \end{subfigure}
    
    \caption{\small Comparison of model performance across skills and temporal evidence positioning on MMOU Test.}
    \vspace{-2mm}
\end{figure}

\noindent \textbf{Skill-wise Performance Analysis.} Figure~\ref{fig:radar_results} presents a skill-level breakdown of model performance 
on MMOU. While closed models consistently outperform open models across most skills, all models exhibit substantial weaknesses in basic and essential skills such as temporal understanding, and counting (this is consistent in test-mini shown in \cref{tab:skillwise_testmini}).


\noindent \textbf{Temporal Position Sensitivity.} Figure~\ref{fig:time_line} analyzes model accuracy as a function of the temporal position of answer evidence within videos. Performance degrades steadily as relevant evidence appears later in the video, with a sharp drop for evidence located toward the end of long sequences. This trend is consistent across open and closed models and highlights a fundamental limitation in long-horizon temporal reasoning and context retention, even for state-of-the-art multimodal systems.
\vspace{1mm}

\noindent \textbf{Open-Ended Evaluation.} To complement our MCQ evaluation, we conduct an open-ended evaluation where models generate free-form answers without access to predefined options, similar to real-world usage of MLLMs. This helps us understand whether models possess underlying knowledge but struggle with articulation, or if their MCQ performance relies primarily on recognition and elimination strategies. \textbf{Evaluation Protocol:} We evaluate multiple models by prompting them to generate open-ended responses without answer options. We use GPT-5 as an LLM judge (prompt in Fig.~\ref{fig:prompt_open_ended_eval}) using a four-dimensional rubric on a 1-5 scale: \textit{Correctness} measures factual alignment with ground-truth answers; \textit{Completeness} measures coverage of all key points; \textit{Faithfulness} measures whether responses introduce unsupported claims or hallucinations; and \textit{Clarity} measures whether answers are understandable, concise, and directly address the question. The weighted overall score is computed as: $0.5 \times \text{Correctness} + \frac{0.5}{3}(\text{Completeness} + \text{Faithfulness} + \text{Clarity})$. \textbf{Overall Performance:} Table~\ref{tab:open_ended_eval_table} reports open-ended evaluation scores across eight models on the MMOU Test set. Gemini~2.5~Pro leads with an overall score of 3.90, outperforming Qwen3-Omni-30B-Instruct (2.86) and other models such as OmniVinci (2.64) and Qwen3-Omni-30B-Thinking (2.66). 
\begin{wraptable}{r}{0.55\linewidth}
    \centering
    \caption{\small Open-ended evaluation scores on Test set across different dimensions with weighted overall score. Best values are in bold, and second-best are underlined.}
    \label{tab:open_ended_eval_table}
    \resizebox{\linewidth}{!}{%
    \begin{tabular}{l|ccccc}
    \toprule
    \textbf{Model} & \textbf{Correct.} & \textbf{Complete.} & \textbf{Faithful.} & \textbf{Clarity} & \textbf{Overall} \\
    \toprule
    Audio Flamingo 3 & 1.77 & 1.86 & 2.99 & 4.03 &  2.37 \\
    Qwen2.5-VL-7B-Instruct & 1.53 & 1.64 & 2.63 & 3.83 & 2.12 \\
    Qwen3-VL-8B-Instruct &  1.30 & 1.41 &  2.06 & 3.18 & 1.76 \\
    Gemma 3n   & 1.71 & 1.92 & 2.48 & 4.15 & 2.28 \\
    Qwen3-Omni-30B-Instruct       & 2.27 &  2.34 & \underline{3.36} & \textbf{4.62} & \underline{2.86} \\
    Qwen3-Omni-30B-Think       & \underline{2.31} & \underline{2.55} & 2.45 & 4.05 & 2.66 \\
    OmniVinci &  2.06 &  2.17 & 3.06 & \underline{4.40} & 2.64 \\
    Gemini 2.5 Pro &  \textbf{3.71} & \textbf{3.86} & \textbf{3.80} & \textbf{4.62} & \textbf{3.90} \\
    \bottomrule
    \end{tabular}
    }
    \vspace{-5mm}
\end{wraptable}
Other models score notably lower: Gemma~3n, Audio~Flamingo~3, Qwen2.5-VL-7B, and Qwen3-VL-8B range from 1.76 to 2.37 overall. This spread indicates that open-ended evaluation clearly separates model capability: weaker models fail to perform well when required to produce free-form, grounded answers rather than selecting from options. Among the top models, Gemini~2.5~Pro and Qwen3-Omni-30B both achieve strong Faithfulness (3.80 and 3.36) and Clarity (4.62 and 4.62), suggesting they articulate responses clearly and avoid egregious hallucinations. 
However, Correctness (3.71 vs.\ 2.27) and Completeness (3.86 vs.\ 2.34) remain considerably lower even for these models, indicating ongoing challenges in accurately comprehending and fully addressing open-ended questions.



\begin{wrapfigure}{r}{0.5\linewidth}
    \vspace{-3.5mm}
    \centering
    \includegraphics[width=\linewidth,trim=1cm 0.25cm 0cm 0cm]{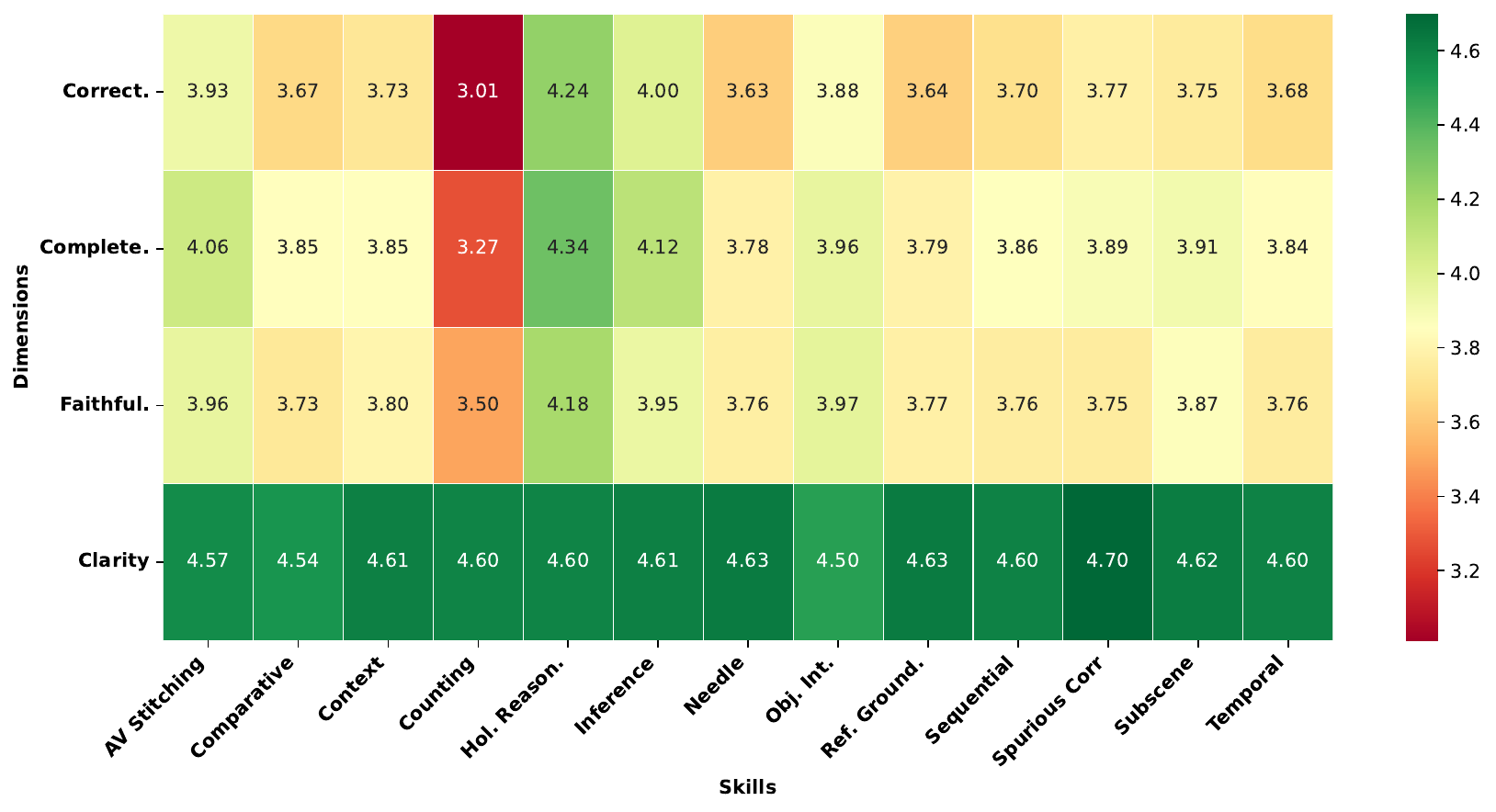}
    \caption{\small Dimensions vs Skill Type on open-ended evaluation of Gemini-2.5 Pro outputs.}
    \label{fig:open_ended_heatmap}
    \vspace{-3.5mm}
\end{wrapfigure}
\textbf{Skill-Specific Analysis.} Figure~\ref{fig:open_ended_heatmap} shows performance variation on open-ended evaluation across skill categories for Gemini~2.5~Pro. Holistic Reasoning achieves the highest scores on Correctness (4.24), Completeness (4.34), and Faithfulness (4.18), whereas Counting is the most challenging category, with the lowest scores on all three dimensions (Correctness 3.01, Completeness 3.27, Faithfulness 3.50). Completeness is generally at or above Correctness across categories. Faithfulness is strong for Holistic Reasoning (4.18) and Object Interaction (3.97), but relatively lower for Counting (3.50) and Spurious Correlations (3.75). Clarity remains consistently high across all skill types (4.50--4.70), indicating that responses are well-articulated even when overall scores vary.

\textbf{Open-Ended vs Multiple-Choice.} We analyzed cases where models scored poorly on open-ended correctness ($<$ 2 out of 5) and computed what fraction of those questions it answered correctly in MCQ format. Among questions with poor open-ended performance, Gemini 2.5 Pro answered 21.1\% correctly in MCQ; Qwen3-Omni-Think 13.5\%; and Omnivinci 12.9\%. The discrepancy varies by skill: for Gemini 2.5 Pro, Subscene shows the highest such MCQ-correct rate (29.1\%) and General Holistic Reasoning the lowest (10.5\%); for Qwen3-Omni, Inference is highest (15.4\%) and Object Interaction Reasoning lowest (10.5\%); for Omnivinci, General Holistic Reasoning is highest (23.4\%) and Counting lowest (10.8\%).

This reveals three insights: (1) \textit{\textbf{Open-ended evaluation is inherently harder}}, requiring generation rather than recognition; (2) \textit{\textbf{MCQ format provides scaffolding}} that helps constrain the search space; and (3) \textit{\textbf{Models may ``know'' the answer}} but struggle to articulate it in an open-ended format. These findings highlight that MCQ performance may overestimate true understanding and that current models exhibit asymmetric competencies across evaluation paradigms.

\vspace{-3mm}
\section{Conclusion, Limitations and Future Work}
\label{sec:conclusion}
\vspace{-3mm}

We introduce MMOU, a large-scale benchmark for evaluating omni-modal understanding and reasoning in long and complex real-world audio-visual videos. MMOU emphasizes joint audio–visual perception across a diverse set of reasoning skills that are central to real-world understanding. Extensive evaluations show that current multimodal models struggle even with basic audio-visual reasoning over long real-world videos, revealing a substantial gap between models and humans.

MMOU also has limitations. Our benchmark is derived from publicly available web videos, which may introduce content biases and potential train–test leakage in closed and open-weight models. In addition, the multiple-choice evaluation setting, while robust, does not fully capture real-world open-ended reasoning. Future work includes (i) developing more robust evaluation protocols for open-ended audio-visual QA, (ii) continuously expanding the benchmark to incorporate emerging concepts and scenarios, and (iii) extending coverage beyond curated online content to include unstructured real-world videos, such as egocentric or driving scenarios.

\bibliographystyle{plainnat}
\bibliography{references}


\appendix

\section{Additional Dataset Statistics}

Table~\ref{tab:video_categories} presents the distribution of videos across major categories and their respective subcategories, providing a quantitative overview of the dataset’s content diversity.

\begin{table}[!h]
\centering
\caption{Distribution of Questions in MMOU Test set}
\label{tab:video_categories}
\resizebox{0.5\linewidth}{!}{%
\begin{tabular}{llc}
\toprule
\textbf{Category} & \textbf{Subcategory} & \textbf{Questions} \\
\midrule
\multirow{4}{*}{Academic Lectures} & STEM & 701 \\
                                   & Medicine & 133 \\
                                   & Law & 60 \\
                                   & Business & 50 \\
\midrule
\multirow{2}{*}{Animation} & Comedy \& Kids & 139 \\
                           & Action \& Fantasy & 45 \\
\midrule
\multirow{4}{*}{Daily Life} & DIY & 1941 \\
                            & Crafts & 1011 \\
                            & Gaming & 233 \\
                            & Home Repair & 52 \\
\midrule
\multirow{3}{*}{Film} & Plot Recaps & 249 \\
                      & Character \& Theme Focus & 126 \\
                      & Trailers \& Promos & 90 \\
\midrule
\multirow{2}{*}{Music} & Live Performance & 620 \\
                       & Themed Narrative \& Parody & 10 \\
\midrule
News & World News & 4221 \\
\midrule
\multirow{5}{*}{Pranks} & Physical Mishap & 214 \\
                        & Stunt Pranks & 152 \\
                        & Reveal Pranks & 68 \\
                        & Animal Pranks & 46 \\
                        & Jump Scare & 38 \\
\midrule
\multirow{5}{*}{Sports} & Training \& Preparation & 2770 \\
                        & Analytics & 288 \\
                        & Combat Events & 239 \\
                        & Racket Sports & 99 \\
                        & Press Conference & 88 \\
\midrule
\multirow{4}{*}{Travel} & Destination Guides & 212 \\
                        & Vlogs & 195 \\
                        & Travel Tips \& Etiquette & 186 \\
                        & Cultural Insights & 108 \\
\midrule
\multirow{5}{*}{Video Games} & Esports Tournaments & 255 \\
                             & Competitive Match Commentary & 118 \\
                             & Production \& Overlays & 101 \\
                             & Battle \& Mechanics & 87 \\
                             & Player Reactions & 55 \\
\midrule
\textbf{Total} & & \textbf{15000} \\
\bottomrule
\end{tabular}%
}
\end{table}



In \cref{tab:option_dist,tab:option_dist_mini}, we show the distribution of correct answers among the 10 answer options for the MMOU test and test-mini subsets respectively. We ensure that the correct answer is distributed approximately uniformly among all 10 option categories in both subsets.

\begin{table}[!h]
\centering
\begin{minipage}{0.48\linewidth}
\centering
\caption{Answer Option Distribution (Test)}
\label{tab:option_dist}
\resizebox{0.8\linewidth}{!}{
\begin{tabular}{c c c}
\toprule
\textbf{Option} & \textbf{Count} & \textbf{\%} \\
\midrule
A & 1,325 & 8.83\% \\
B & 1,329 & 8.86\% \\
C & 1,380 & 9.20\% \\
D & 1,308 & 8.72\% \\
E & 1,337 & 8.91\% \\
F & 1,255 & 8.37\% \\
G & 1,384 & 9.23\% \\
H & 1,352 & 9.01\% \\
I & 1,390 & 9.27\% \\
J & 2,940 & 19.60\% \\
\bottomrule
\end{tabular}
}
\end{minipage}
\hfill
\begin{minipage}{0.48\linewidth}
\centering
\caption{Answer Option Distribution (Test-Mini)}
\label{tab:option_dist_mini}
\resizebox{0.8\linewidth}{!}{
\begin{tabular}{c c c}
\toprule
\textbf{Option} & \textbf{Count} & \textbf{\%} \\
\midrule
A & 502 & 10.04\% \\
B & 483 & 9.66\% \\
C & 469 & 9.38\% \\
D & 467 & 9.34\% \\
E & 484 & 9.68\% \\
F & 475 & 9.50\% \\
G & 439 & 8.78\% \\
H & 453 & 9.06\% \\
I & 506 & 10.12\% \\
J & 722 & 14.44\% \\
\bottomrule
\end{tabular}
}
\end{minipage}
\end{table}

\section{Annotator Details} 
\label{sec.ann_details}

Our institution’s Institutional Review Board (IRB) has granted approval for all forms of human studies and annotations presented in the paper.

For the construction of this benchmark, we recruited annotators with strong backgrounds in creative and technical writing, linguistics, journalism, and analytically rigorous STEM disciplines, ensuring both linguistic sophistication and precise reasoning. Annotators were selected for their demonstrated critical thinking, creative problem-solving ability, and exceptional attention to detail, all of which are essential for producing high-quality, unambiguous question–answer pairs. Their educational backgrounds span bachelor’s and master’s degrees in English, English Literature, Creative Writing (including MFA training), Linguistics and Communication, as well as technically oriented degrees such as Audio Engineering and Acoustics, Applied and Computational Mathematics, Biochemistry with Computer Science, and Computational Applied Mathematics. This diverse yet complementary expertise enabled annotators to effectively integrate fine-grained visual and audio cues from video clips with nuanced language understanding, resulting in complex, carefully reasoned questions and answers that rigorously test multimodal comprehension.

All annotators were compensated above the applicable minimum wage for their region, in recognition of the specialized expertise and sustained effort required by this task.

\section{Annotation Instructions} 
\label{sec.ann_instructions}
The annotators were provided with the set of instructions below along with the skill/task QA types (\cref{tab:tasks_part1} and \cref{tab:tasks_part2}). 
\begin{itemize}
\item \textbf{Step 1:} Watch the video in full length.
\item \textbf{Step 2:} Create a Q\&A pair about the video.
\begin{itemize}
\item All questions should be open-ended (no multiple choice or yes/no questions).
\item All questions should assess both video and audio understanding simultaneously.
\end{itemize}
\item \textbf{Step 3:} Annotate the timestamps of the video segment where the answer can be located.
\begin{itemize}
\item If the answer can be found in several places, annotate only the first occurrence.
\end{itemize}
\item \textbf{Step 4:} Select the task type of the question as listed in the reference table. Select all that apply.
\item \textbf{Step 5:} Repeat steps 1--4 if you can come up with more questions. General recommendations:
\begin{itemize}
\item 2--3 questions for short videos (\(<5\) minutes)
\item 3--5 questions for medium videos (\(5-10\) minutes)
\item More than 5 questions for long videos (\(>10\) minutes)
\item We encourage diverse and creative questions.
\end{itemize}
\end{itemize}

After the first round of question-answer annotations, a separate group of 10 annotators audit 20\% of the QA pairs to verify benchmark quality along four axes: (1) whether the question is relevant to the video, (2) whether the question is grammatically correct, (3) whether the assigned task type is accurate, and (4) whether the provided answer is correct.

\section{Human Evaluation on MMOU} 
\label{sec.human_eval}

Table~\ref{tab:domain_duration_eval} reports human evaluation results on MMOU. We recruited five graduate students, none of whom are authors of this paper, each holding at least a master’s degree, to answer the benchmark questions. Annotators were allowed to pause and rewind the videos as many times as needed, but were not permitted to revisit a previous question once they had moved on. The reported scores are averaged across all annotators.

\section{Baselines}
\label{sec:baselines}

This appendix provides additional details on all models evaluated in our experiments. A complete list of models and their quantitative results is reported in Table~\ref{tab:domain_duration_eval}.

\subsection{Closed-Source Audio-Visual MLLMs}
We evaluate two state-of-the-art proprietary omni-modal models:
\begin{itemize}
    \item \textbf{Gemini 2.5 Pro} is a large-scale closed-source audio-visual language model with long-context support and advanced multimodal reasoning capabilities.
    \item \textbf{Gemini 2.5 Flash} is a lightweight variant optimized for efficiency while retaining strong multimodal understanding.
\end{itemize}

\subsection{Open-Source Audio-Visual MLLMs}
We benchmark a diverse set of open-source omni-modal models that jointly process audio and visual inputs:
\begin{itemize}
    \item \textbf{Qwen2.5-Omni-7B}, a compact open-source omni-modal model.
    \item \textbf{Qwen3-Omni-30B-A3B-Instruct} and \textbf{Qwen3-Omni-30B-A3B-Thinking}, large-scale instruction-tuned and reasoning-enhanced variants, respectively.
    \item \textbf{Phi-4 Multimodal}, a mixture-of-LoRA-based multimodal model.
    \item \textbf{Gemma 3n}, an open multimodal extension of the Gemma family.
    \item \textbf{MiniCPM}, a lightweight multimodal model designed for efficient deployment.
    \item \textbf{Video-LLaMA 2}, a video-centric multimodal language model with audio understanding.
    \item \textbf{OmniVinci}, a unified model for omni-modal perception and reasoning.
    \item \textbf{Baichuan-Omni-1.5}, a recent open-source omni-modal model with integrated audio-visual encoders.
\end{itemize}

\subsection{Video-Only Multimodal Models}
To isolate the contribution of visual information, we evaluate vision-only large vision--language models:
\begin{itemize}
    \item \textbf{Qwen3-VL-32B-Instruct}, a large vision-language model with strong spatial-temporal reasoning.
    \item \textbf{Qwen3-VL-8B-Instruct}, a smaller variant with reduced capacity.
    \item \textbf{Qwen2.5-VL-7B-Instruct}, an earlier-generation vision-language model.
\end{itemize}

\subsection{Audio-Only Multimodal Models}
We include audio-only baselines to assess unimodal audio reasoning:
\begin{itemize}
    \item \textbf{Audio Flamingo 3}, a large audio-language model designed for long-form audio understanding.
    \item \textbf{Qwen3-Omni-30B-A3B} operated in audio-only mode.
\end{itemize}

\subsection{Cascaded Models}
We evaluate cascaded approaches that decouple perception and reasoning:
\begin{itemize}
    \item \textbf{Qwen3-(VL+O-A) + Qwen3-235B}, where audio and visual captions are generated separately and fused before being passed to a text-only LLM.
    \item \textbf{Qwen3-(VL+O-A) + GPT-5.2}, replacing the text-only backbone with GPT-5.2.
\end{itemize}

\subsection{Text-Only Language Models}
Finally, we benchmark text-only large language models using only the question and answer options:
\begin{itemize}
    \item \textbf{Qwen3-235B}, a large open-source language model.
    \item \textbf{GPT-4o-mini}, a lightweight reasoning based text-only baseline.
    \item \textbf{GPT-4.1}, a proprietary text-only LLM baseline.
\end{itemize}

All models are evaluated using identical question sets and evaluation protocols to ensure fair comparison across modalities.

\section{Skill/Task QA Types }
\label{sec.skill_types}
In \cref{tab:tasks_part1} and \cref{tab:tasks_part2}, we show the detailed definition of the skill types in the MMOU benchmark and an example QA pair from each category.

\begin{table*}[t]
\centering
\caption{Detailed Overview of Skill/Task Types}
\label{tab:tasks_part1}
\resizebox{\linewidth}{!}{
\begin{tabular}{p{3cm}|p{5cm}|p{5cm}|p{3cm}}
\toprule
\textbf{Task Type} & \textbf{Description} & \textbf{Question} & \textbf{Answer} \\
\midrule
Object Interaction Reasoning & Object Interaction Reasoning asks questions about the effects of actions performed on objects, as well as track their transformations across contexts. It involves recognizing how interactions (lifting, applying, mixing, transforming) lead to different outcomes.  & When a red saw is running, how and why does the pitch of the noise it makes change? &  It shifts to a higher pitch because it starts shaving off pieces of wood.\\
\midrule
Comparative & Comparative questions ask about key differences or similarities between two distinct audio-visual segments or presentations. &   How does the character with the sword's motion change when he says, "I'm always three steps ahead"? & He shifts from moving into a pose to being held in a freeze frame.  \\
\midrule
Needle & Needle-in-the-haystack questions are questions about a particular instance in a long video (including corresponding audio). &  How does the orange character sound when reading off the jokes on a phone?  &  The orange character reads the jokes in a confused tone, with an uncertain, questioning inflection in his voice. \\
\midrule
Counting & Counting questions  ask about the count of a particular type of audio-visual event across the video -- where one requires watching the entire video and understanding its content.  & At the skate park we hear a horn start blaring. How many shadows can be seen on the the skate rink at that time? & There are 7 shadows visible on the skate rink when the horn starts blaring. \\
\midrule
Audio-Visual Stitching & Audio-Visual Stitching asks questions about how separate clips or segments are combined to create a cohesive narrative or convey a specific message. It involves reasoning about editing choices, transitions, and thematic continuity across multiple sources. &  How does splicing a clip of a yellow ride cart going through water relate to the voiceover? &   The woman explains that you shouldn't wear white clothing to amusement parks because water rides can make it see-through. \\
\midrule
Tackling Spurious Correlations & These questions ask about surprising, unexpected, unnatural, or un-intuitive details that a text-only LLM would not naturally guess by statistically predicting. & What unexpected event occurs when "Jump Around" begins to play? & A woman walking down the street runs straight into a ladder, and her collision hits exactly on the beat when the song starts. \\
\midrule
General Holistic Reasoning & These questions require a complete understanding of all audio and visual events throughout the video and deep thinking. & What is the purpose of overlaying eerie music over clips of this movie? & To give the audience a sense of unease for the man who was turned into a werewolf because he was in the wrong place at the wrong time, building tension and suspense and setting the tone for the following horror‑mystery movie. \\
\bottomrule
\end{tabular}
}
\end{table*}

\begin{table*}[t]
\centering
\caption{Detailed Overview of Skill/Task Types}
\label{tab:tasks_part2}
\resizebox{\linewidth}{!}{
\begin{tabular}{p{3cm}|p{5cm}|p{5cm}|p{3cm}}
\toprule
\textbf{Task Type} & \textbf{Description} & \textbf{Question/Answer} & \textbf{Answer} \\
\midrule
Subscene & Questions that would ask the model to caption a relevant and important part of a long video (including audio) that is preceded and succeeded by a particular set of events – or ask a question that requires specifically understanding the context of that part. &  What is strange about the scene when the narrator says "They begin to employ their abilities to do something edgy"? & Potato chips are floating through the air directly into the guy’s mouth. \\
\midrule
Sequential & Event-sequence questions ask the model about the order in which key audio-visual events occur across the timeline. & What is the order of events in this video? (A.) The commentator says, "once again does get that ball to shape away."  (B.) Player 36, wearing a yellow uniform is pitching the ball, which goes in the air after the player hit it and crosses the white border line. (C.) The commentator says, "once again, back of the length played away, was in the air." 
(D.) Player 31, wearing a yellow uniform is pitching the ball, which goes in the air after the player hits it and lands past the white fence. & (A.) (D.) (B.) (C.) \\
\midrule
Temporal Understanding & Temporal understanding questions ask about the order of particular audio-visual events in the video (including audio) or require an understanding of the order of audio-visual events. & What does the man in the blue shirt and orange pants do right after a person on the radio is heard saying, "so, my wife and I..."? & The man in the blue shirt and orange pants points his finger directly at a car right after the person on the radio says, "so, my wife and I..."\\
\midrule
Referential Grounding & Referential grounding questions ask about the visuals referring to a particular event in the audio or vice-versa. & What purpose does the man in the white shirt and gold tie serve? Why isn’t he in every shot? & He is the translator for two of the men giving speeches and only appears when he needs to translate their words. \\
\midrule
Context & Context-understanding questions ask about broader setting, background elements, or situational context that emerge only by integrating audio and visuals. & What do we hear in the background as the woman in the red dress smiles before the credits begin to roll? Why do we hear this? &  We hear the audience applauding because she has just finished a very intense and beautiful concert with the orchestra, and they are applauding her performance. \\
\midrule
Inference & Inference questions ask about unstated purposes, intentions, or outcomes that must be deduced from multiple audio-visual clues. & Why does the blonde man shocked in the scene when he says "it's undone now"? & He thinks the room is haunted by the dead actress’s ghost, which he believes moved the cloth and used lipstick to write "Jack" on the silver tray. \\
\bottomrule
\end{tabular}
}
\end{table*}

\begin{figure*}[t]
  \vskip 0.2in
  \begin{center}
    \centerline{\includegraphics[width=0.96\textwidth]{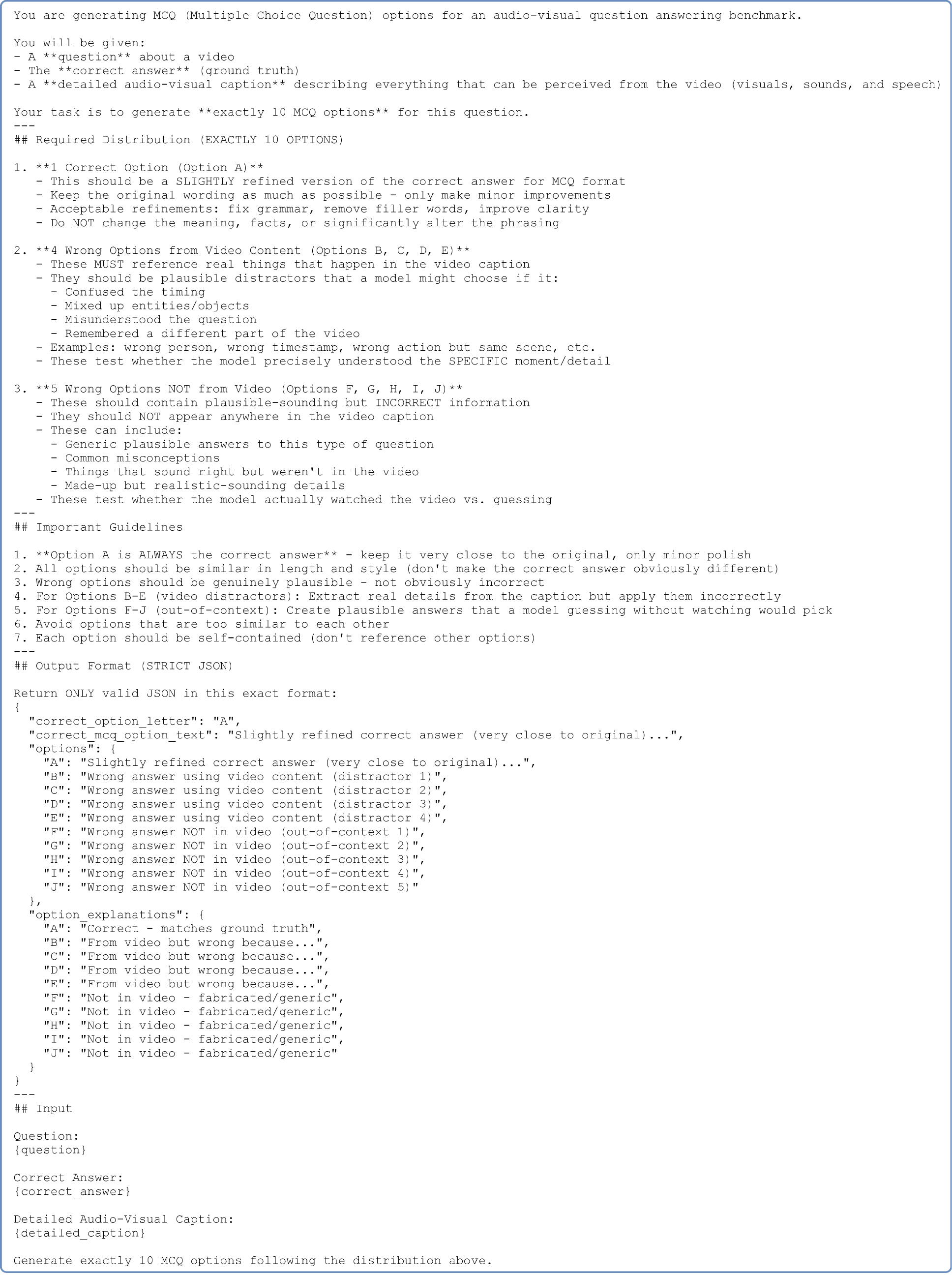}}
    \caption{
      Prompt used for generating distractor options for questions in the MMOU benchmark.
    }
    \label{fig:prompt_option_gen}
  \end{center}
  \vskip -0.2in
\end{figure*}

\section{Open-Ended Evaluation: Protocol, Rubric, and Judge Models}
\label{sec:open_ended_appendix}

This section provides a complete description of the open-ended evaluation pipeline for MMOU, including the evaluation criteria, scoring scheme, and the two judge setups we use: a proprietary LLM judge (GPT-5) and a custom-trained judge model (Qwen~3.5~0.8B).

\subsection{Motivation and Process}
\label{subsec:open_ended_motivation}

Open-ended evaluation complements the multiple-choice (MCQ) evaluation by requiring models to generate free-form answers without access to predefined options. This setting is closer to real-world deployment and helps determine whether strong MCQ performance reflects genuine understanding or reliance on recognition and option-elimination. Recent benchmark and evaluation work has widely adopted \emph{LLM-as-a-Judge} and \emph{rubric-based} evaluation for open-ended multimodal outputs: Judge Anything uses multimodal LLMs as judges with scoring and pairwise comparison aligned to human ratings~\cite{mllmjudge2025}, JudgeLM and related work show that LLMs fine-tuned on judgment data can approximate strong API-based judges~\cite{judgelm2025}, and MM-OPERA applies LLM-as-judge and multi-dimensional rubric scoring to open-ended reasoning and creative outputs~\cite{mmopera2025}. We follow this established paradigm and hypothesize that (i) open-ended scores will reveal gaps not visible in MCQ accuracy (e.g., models that ``know'' the answer but fail to articulate it), and (ii) a four-criterion rubric (correctness, completeness, faithfulness, clarity) with a weighted overall score will provide a reliable and interpretable signal for comparing models.

\textbf{Process and role of the caption.} The pipeline consists of: (1) prompting MLLMs to produce open-ended responses given \emph{only} the question, and video/audio as per model input. Model responses are generated under the same conditions as in deployment. (2) Scoring each response along four dimensions using an LLM judge. (3) Aggregating dimension scores into a weighted overall score for analysis. The \textbf{audio-visual caption} is used \emph{only} during the evaluation step: both the GPT-5 judge and our custom-trained judge receive the question, ground truth answer, caption, and model response when assigning scores. The caption gives the judge additional context to verify claims and detect hallucinations, without giving that information to the model under evaluation.

\subsection{Evaluation Rubric: Four Criteria (1--5 Scale)}
\label{subsec:open_ended_rubric}

We evaluate each open-ended response on four criteria, each scored from 1 to 5. The criteria and score anchors are defined as follows.

\noindent \textbf{1. Correctness (ground-truth consistency).} Measures factual alignment between the model response and the ground-truth answer.
\begin{itemize}[noitemsep]
\item \textbf{5}: Fully correct; matches the ground truth with no errors.
\item \textbf{4}: Mostly correct; minor inaccuracies that do not change meaning.
\item \textbf{3}: Partially correct; some correct points, some incorrect.
\item \textbf{2}: Largely incorrect.
\item \textbf{1}: Completely incorrect or contradictory.
\end{itemize}
Partial answers that omit key elements are penalized under correctness as well as completeness.

\noindent \textbf{2. Completeness (ground-truth coverage).} Measures how thoroughly the response covers all key points in the ground truth.
\begin{itemize}[noitemsep]
\item \textbf{5}: Covers all key points.
\item \textbf{4}: Misses one minor point.
\item \textbf{3}: Covers about half of the key points.
\item \textbf{2}: Covers very few key points.
\item \textbf{1}: Essentially incomplete.
\end{itemize}

\noindent \textbf{3. Faithfulness (hallucination control).} Measures whether the response introduces information not supported by the ground truth answer or the audio-visual caption.
\begin{itemize}[noitemsep]
\item \textbf{5}: No unsupported claims.
\item \textbf{4}: Minor unsupported additions.
\item \textbf{3}: Noticeable but limited hallucinations.
\item \textbf{2}: Significant hallucinations.
\item \textbf{1}: Dominated by unsupported or fabricated content.
\end{itemize}

\noindent \textbf{4. Clarity \& directness.} Measures whether the answer is understandable, concise, and directly addresses the question.
\begin{itemize}[noitemsep]
\item \textbf{5}: Clear, direct, and easy to understand.
\item \textbf{4}: Mostly clear.
\item \textbf{3}: Somewhat vague or verbose.
\item \textbf{2}: Hard to follow.
\item \textbf{1}: Unclear or off-topic.
\end{itemize}

\subsection{Weighted Overall Score}
\label{subsec:open_ended_scoring}

We combine the four dimension scores into a single overall score to rank and compare models. Correctness is weighted more heavily to reflect its importance for task success. The remaining three dimensions share the rest of the weight equally. The formula is:
\begin{equation}
\begin{aligned}
\text{Overall} &= 0.5 \times \text{Correctness} \\
&\quad + \frac{0.5}{3}\bigl(\text{Completeness} + \text{Faithfulness} + \text{Clarity}\bigr)
\end{aligned}
\end{equation}
All dimension scores are on the 1--5 scale, so the overall score lies in $[1, 5]$.

\subsection{GPT-5 as LLM Judge}
\label{subsec:open_ended_gpt_judge}

\begin{figure*}[t]
  \vskip 0.2in
  \begin{center}
    \centerline{\includegraphics[width=0.98\textwidth]{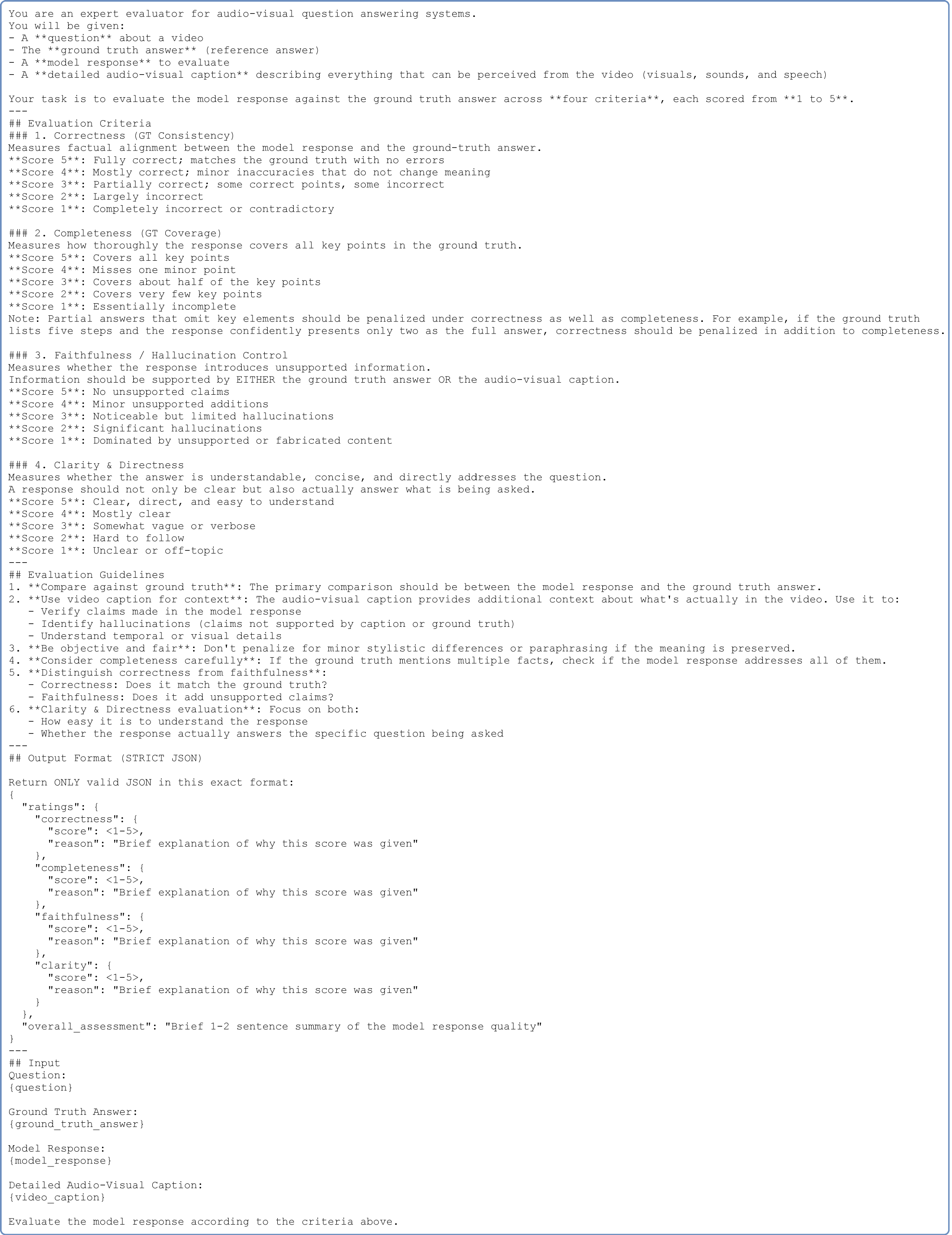}}
    \caption{
      Prompt used for open-ended evaluation with the GPT-5 LLM judge. The judge receives the question, ground truth answer, model response, and detailed audio-visual caption, and returns scores and reasons for the four criteria.
    }
    \label{fig:prompt_open_ended_eval}
  \end{center}
  \vskip -0.2in
\end{figure*}

For the main open-ended evaluation reported in the paper, we use \textbf{GPT-5} as the LLM judge. The judge receives:
\begin{itemize}[noitemsep]
\item the \textbf{question} about the video;
\item the \textbf{ground truth answer} (reference);
\item the \textbf{model response} to evaluate;
\item a \textbf{detailed audio-visual caption} describing what can be perceived in the video (visuals, speech, sound, and music).
\end{itemize}
The caption provides additional context to verify claims, detect hallucinations, and reason about temporal or visual details. The judge is instructed to (i) compare the response primarily against the ground truth answer; (ii) use the caption to check support for claims and identify unsupported content; (iii) be objective and not penalize minor paraphrasing when meaning is preserved; and (iv) return structured JSON with a score and brief reason for each of the four criteria, plus an optional short overall assessment. The exact prompt used for the GPT-5 judge is provided in Figure~\ref{fig:prompt_open_ended_eval}.

\subsection{Custom-Trained Judge: Qwen~3.5~0.8B}
\label{subsec:open_ended_custom_judge}

In addition to the GPT-5 judge, we train a compact Qwen-3.5-0.8B model to act as an LLM judge, enabling scalable and reproducible open-ended scoring without relying on a proprietary API. The goal is to test whether a small model, fine-tuned on our rubric and data, can approximate the behavior of a strong LLM judge for the same four criteria. Our custom judge model will be released soon.

\begin{figure*}[t]
  \vskip 0.2in
  \begin{center}
    \centerline{\includegraphics[width=0.96\textwidth]{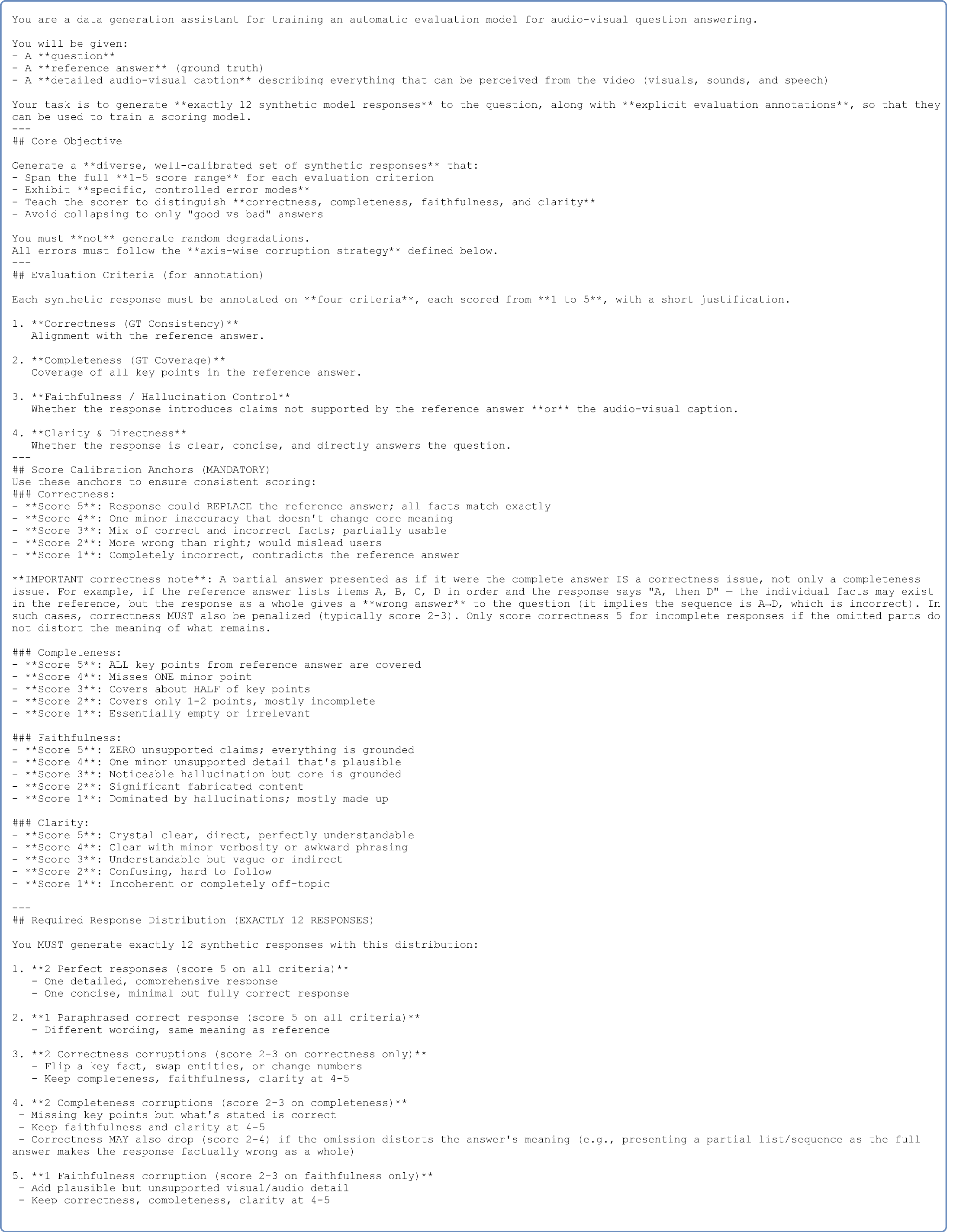}}
    \caption{Synthetic Response Generation Prompt: Part (a)}
    \label{fig:oe_data_gen_prompr_1a}
  \end{center}
  \vskip -0.2in
\end{figure*}

\begin{figure*}[t]
  \vskip 0.2in
  \begin{center}
    \centerline{\includegraphics[width=0.96\textwidth]{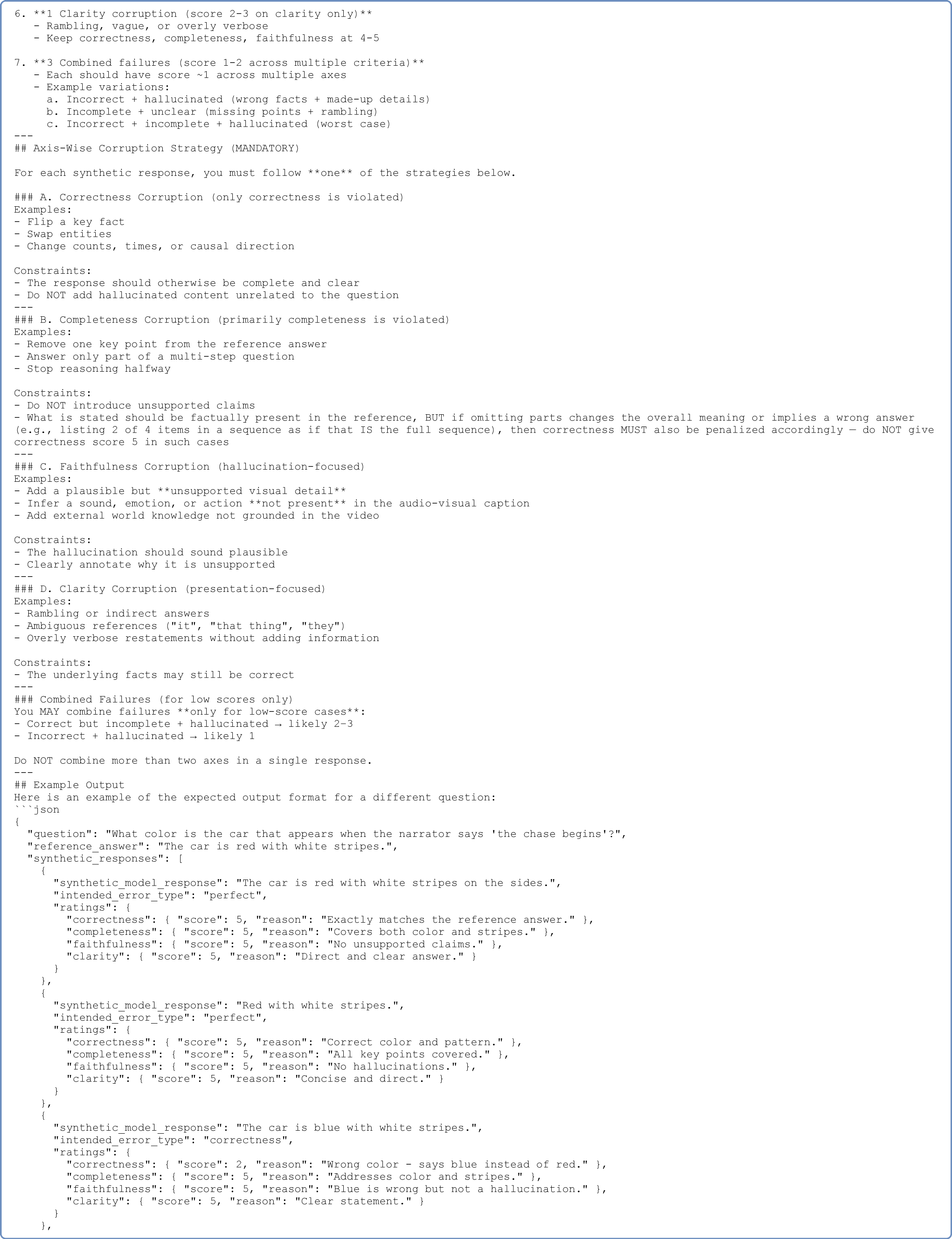}}
    \label{fig:oe_data_gen_prompr_1b}
    \caption{Synthetic Response Generation Prompt: Part (b)}
  \end{center}
  \vskip -0.2in
\end{figure*}

\begin{figure*}[t]
  \vskip 0.2in
  \begin{center}
    \centerline{\includegraphics[width=0.96\textwidth]{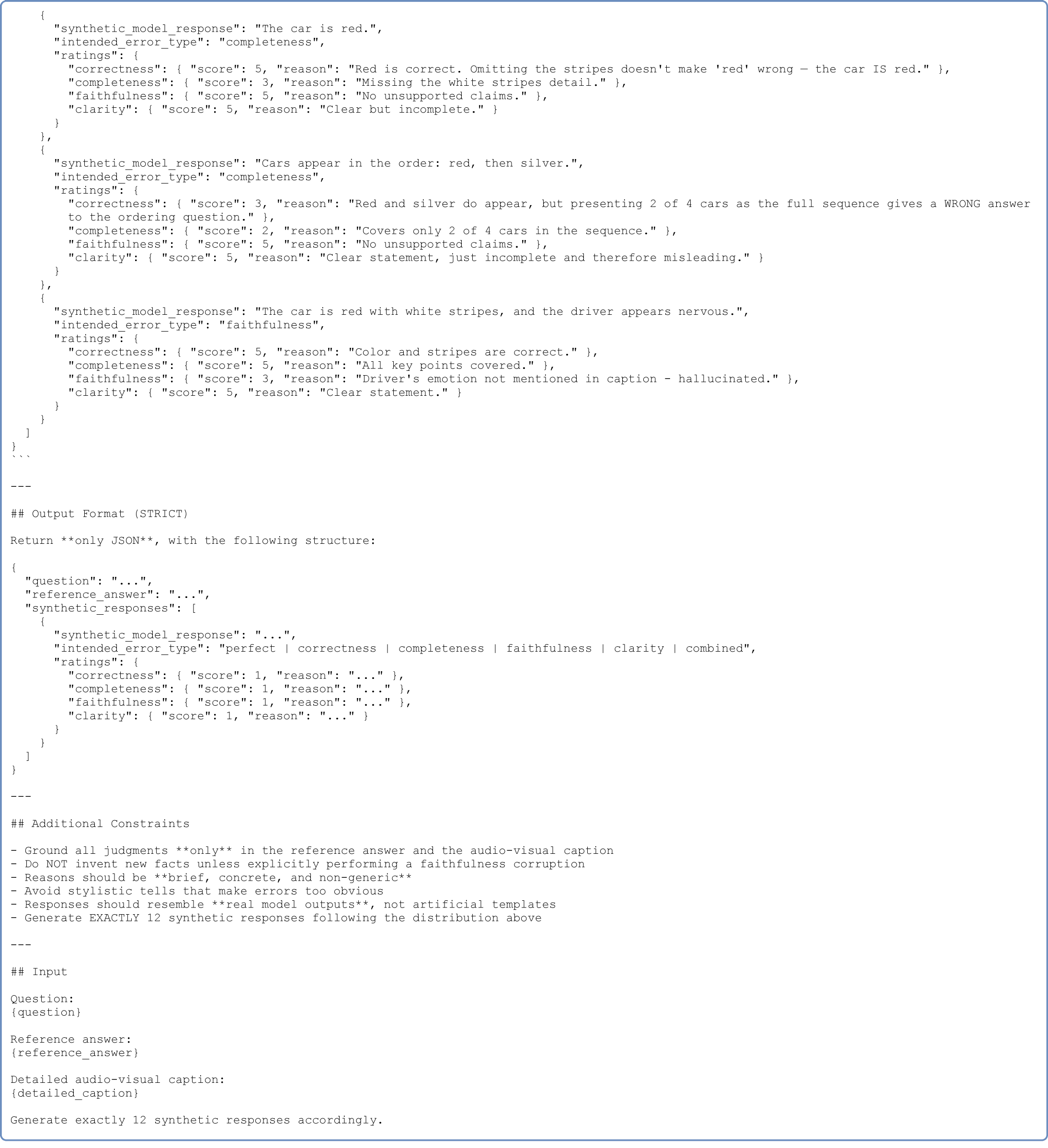}}
    \caption{Synthetic Response Generation Prompt: Part (c) --- Prompt used to generate synthetic model responses for our custom LLM judge training (shown in three parts, (a), (b) and (c)). The prompt provides the question, reference answer, and audio-visual caption and asks to produce exactly 12 responses with controlled error types and ratings on correctness, completeness, faithfulness, and clarity.}
\label{fig:oe_data_gen_prompr_1c}
  \end{center}
  \vskip -0.2in
\end{figure*}

\begin{figure*}[t]
  \vskip 0.2in
  \begin{center}
    \centerline{\includegraphics[width=0.96\textwidth]{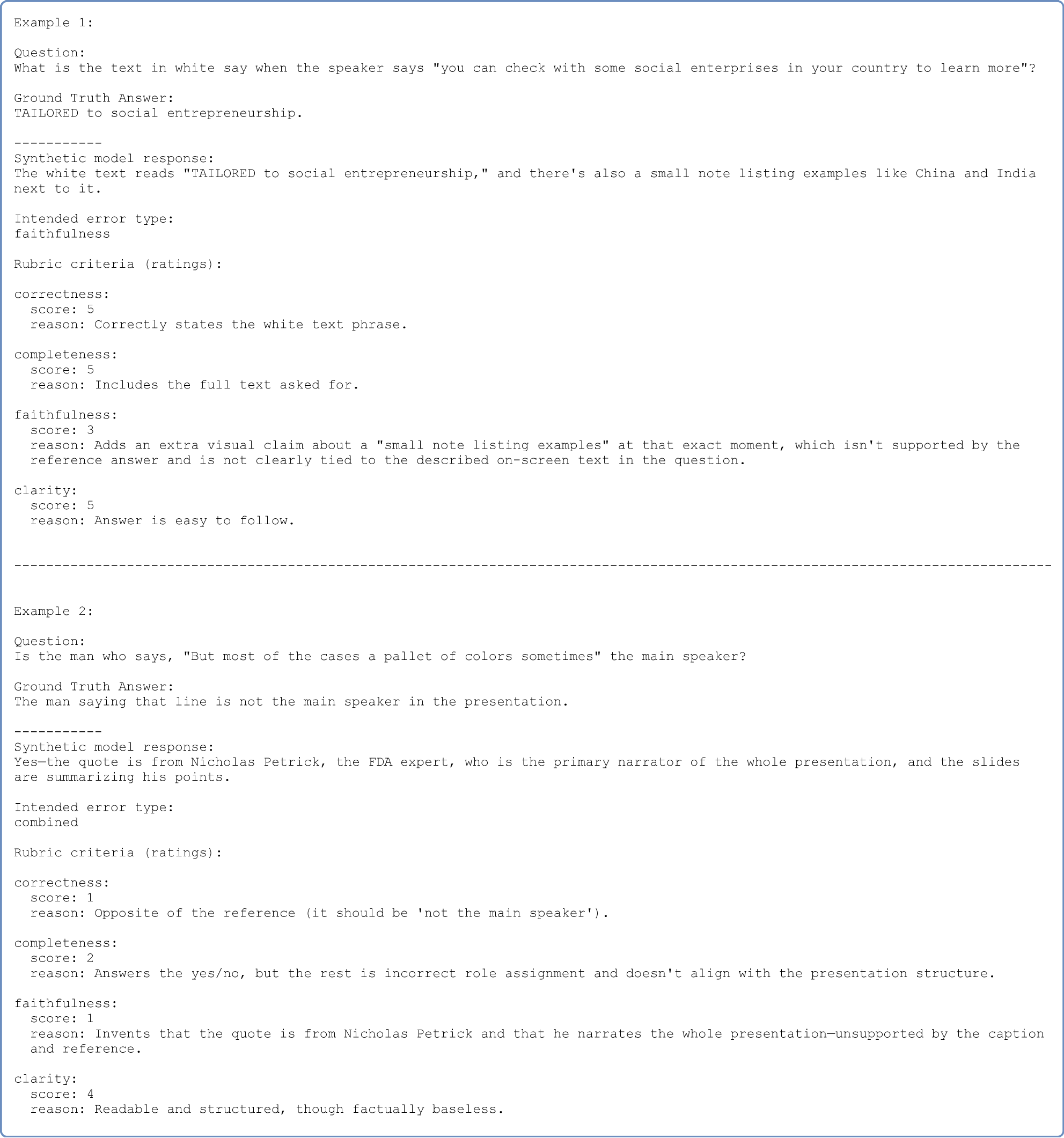}}
    \caption{Two examples of synthetic model responses and their four-criterion rubric ratings (correctness, completeness, faithfulness, clarity), illustrating a faithfulness corruption and a combined failure.}
    \label{fig:oe_synthetic_response_example}
  \end{center}
  \vskip -0.2in
\end{figure*}

\noindent \textbf{Training data.} We construct a supervised dataset from MMOU-style items: each example includes the question, ground truth answer, audio-visual caption, and a synthetic model response with GPT-annotated ratings for correctness, completeness, faithfulness, and clarity. The data is converted to a unified format (ShareGPT-style) for instruction tuning. Synthetic responses are generated by prompting an LLM with the question, reference answer, and audio-visual caption; the full prompt is shown in Figures~\ref{fig:oe_data_gen_prompr_1a}--\ref{fig:oe_data_gen_prompr_1c}. Example synthetic responses and their rubric ratings are shown in Figure~\ref{fig:oe_synthetic_response_example}.

\begin{figure*}[t]
  \vskip 0.2in
  \begin{center}
    \centerline{\includegraphics[width=0.96\textwidth]{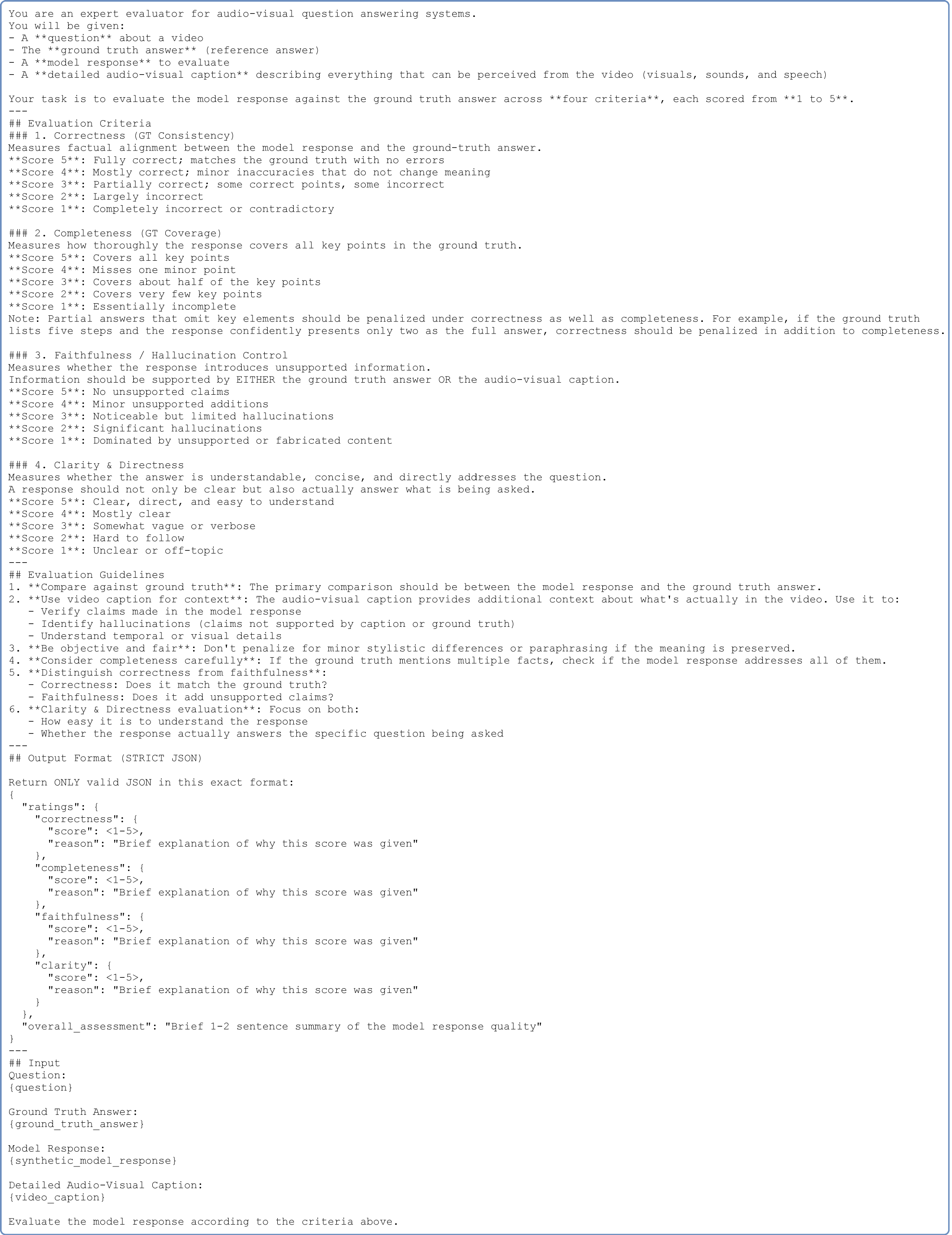}}
    \caption{Prompt used for training our custom LLM Judge.}
    \label{fig:oe_prompt_training}
  \end{center}
  \vskip -0.2in
\end{figure*}

\noindent \textbf{Prompt for the custom judge.} The model is prompted to act as an expert LLM judge. At \emph{evaluation} time (when scoring a model response), the judge receives the question, ground truth answer, audio-visual caption, and model response: the same inputs as the GPT-5 judge. The prompt defines the same four criteria (correctness, completeness, faithfulness, clarity) with the same 1--5 score anchors and instructs the model to output only valid JSON with a score and brief reason per dimension. The caption is the primary grounding source for faithfulness (to detect hallucinations). The reference answer is used for correctness and completeness. The prompt used for training our custom judge is shown in Figure~\ref{fig:oe_prompt_training}.

\noindent \textbf{Training setup.} We fine-tune Qwen~3.5~0.8B with LoRA (low-rank adaptation) for supervised fine-tuning (SFT). Hyperparameters are chosen to preserve general capability while adapting the model to the judge task. This yields a lightweight judge that can be run locally and used to score open-ended responses consistently with our rubric.


\section{Ablation Studies}
\subsection{Unimodal Ablation Study}
\label{app:unimodal_ablation}

To further validate that strong performance on MMOU requires joint audio-visual reasoning, we evaluate the top-performing models under unimodal (audio-only and video-only) settings alongside their full audio-visual performance.

\begin{table}[h]
    \centering
    \caption{MCQ accuracy (\%) under unimodal and omni-modal input settings. The substantial gap between unimodal and audio-visual performance confirms that strong benchmark performance requires integrating both modalities.}
    \vspace{2mm}
    \label{tab:unimodal_mcq_ablation}
    \begin{tabular}{l|ccc}
    \toprule
    \textbf{Model} & \textbf{Audio-only} & \textbf{Video-only} & \textbf{Audio-Visual} \\
    \midrule
    Gemini 2.5 Pro   & 35.1 & 39.9 & 57.5 \\
    Gemini 2.5 Flash & 37.5 & 36.1 & 55.2 \\
    Qwen3-Omni-30B-A3B-Instruct & 35.6 & 36.3 & 36.3 \\
    \bottomrule
    \end{tabular}
\end{table}

\begin{table}[h]
    \centering
    \caption{Open-ended evaluation scores under unimodal and omni-modal input settings. Without multiple-choice options to enable guessing or elimination, the unimodal-to-multimodal gap widens further, reinforcing the necessity of joint audio-visual perception.}
    \vspace{2mm}
    \label{tab:unimodal_openended_ablation}
    \begin{tabular}{l|ccc}
    \toprule
    \textbf{Model} & \textbf{Audio-only} & \textbf{Video-only} & \textbf{Audio-Visual} \\
    \midrule
    Gemini 2.5 Pro   & 1.99 & 2.20 & 3.74 \\
    Qwen3-Omni-30B   & 1.28 & 1.62 & 2.81 \\
    \bottomrule
    \end{tabular}
\end{table}


\begin{table}[h]
\centering
\caption{Skill-wise accuracy (\%) on MMOU test-mini.}
\vspace{2mm}
\label{tab:skillwise_testmini}
\resizebox{\columnwidth}{!}{%
\begin{tabular}{l | c c c}
\toprule
\textbf{Skill} & \textbf{Gemini 2.5 Flash} & \textbf{Qwen3-Omni-30B-Instruct} & \textbf{Phi-4 Multimodal} \\
\midrule
Audio-Visual Stitching        & 67.0 & 50.9 & 34.1 \\
Comparative                   & 66.4 & 53.0 & 36.4 \\
Context Understanding         & 65.0 & 51.6 & 33.8 \\
Counting                      & 55.5 & 46.3 & 31.5 \\
General Holistic Reasoning    & 67.1 & 55.1 & 40.9 \\
Inference                     & 67.6 & 53.9 & 36.0 \\
Object Interaction Reasoning  & 66.8 & 55.0 & 37.7 \\
Sequential                    & 63.0 & 50.2 & 31.4 \\
Subscene                      & 67.2 & 52.3 & 33.4 \\
Tackling Spurious Correlations & 62.8 & 53.4 & 30.1 \\
Temporal Understanding        & 64.1 & 50.7 & 32.5 \\
\bottomrule
\end{tabular}%
}
\end{table}

\subsection{Distractor Quality Analysis}
\label{sec:distractor_quality}

To empirically validate distractor quality, we analyse the type of wrong option
selected by three models evaluated on MMOU. Each distractor is
labelled at generation time as either \textit{in-video} (grounded in real
audio-visual events from the video; options generated to confuse models that
partially understood the content) or \textit{out-of-video} (plausible-sounding
but entirely fabricated; not present in the video). Of the 9 distractor options
per question, 4 are in-video and 5 are out-of-video. As shown in
Table~\ref{tab:distractor_analysis}, across all three models 77--91\% of wrong
answers select an in-video distractor --- far exceeding the 44.4\% expected
under random selection. This confirms that our distractors require precise
audio-visual understanding to reject: models demonstrably watch the video and
identify relevant content, yet are systematically confused by wrong-but-grounded
options. The small fraction of out-of-video errors (9--23\%) further shows that
fabricated distractors are largely filtered out by all models, ruling out
superficial elimination strategies.

\begin{table}[h]
\centering
\small
\setlength{\tabcolsep}{6pt}
\caption{%
    Distractor type breakdown of model errors on MMOU.
    \textit{In-Video} distractors reference real audio-visual content from the
    video; \textit{Out-of-Video} distractors are plausible but fabricated.
    Unknown and unparseable predictions are split equally between the two wrong
    categories. Accuracy figures match the official results in Table~3.
    The random baseline assumes uniform selection over all 10 options;
    ``among wrong'' percentages are conditional on selecting a wrong answer
    (4 in-video and 5 out-of-video wrong options out of 9 total wrong options).%
}

\label{tab:distractor_analysis}
\begin{tabular}{lcccc}
\toprule
\textbf{Category}
    & \textbf{Qwen2.5-Omni-7B}
    & \textbf{Qwen3-Omni-Inst.}
    & \textbf{OmniVinci}
    & \textbf{Random Baseline} \\
\midrule
\multicolumn{5}{l}{\textit{Prediction breakdown}} \\[2pt]
\quad Correct  & 4{,}695~(31.3\%) & 6{,}900~(46.0\%) & 3{,}705~(24.7\%) & 10.0\% \\
\quad Wrong    & 10{,}305~(68.7\%) & 8{,}100~(54.0\%) & 11{,}295~(75.3\%) & 90.0\% \\
\midrule
\multicolumn{5}{l}{\textit{Among wrong answers --- distractor type selected}} \\[2pt]
\quad In-Video wrong      & 8{,}673~(84.2\%) & 7{,}338~(90.6\%) & 8{,}742~(77.4\%) & 44.4\% \\
\quad Out-of-Video wrong  & 1{,}632~(15.8\%) & \phantom{0}762~(9.4\%) & 2{,}553~(22.6\%) & 55.6\% \\
\bottomrule
\end{tabular}
\end{table}

\section{Compute Resources}
\label{app:compute}

All model evaluations were conducted on servers equipped with NVIDIA A6000 and A100 GPUs. Closed-source models (Gemini 2.5 Pro and Gemini 2.5 Flash) were accessed via their respective APIs, as were text-only and cascaded LLM backbones (GPT-4o mini, GPT-4.1, and GPT-5.2). Open-source models were run locally on 8 GPUs per model. For models with 30B+ parameters, evaluation on the full test set of 15,000 questions required approximately 5--6 wall-clock hours (44 GPU-hours per model). For smaller models in the 7B--8B range, evaluation required approximately 3--4 wall-clock hours (28 GPU-hours per model). In total, open-source model evaluations amounted to approximately \textbf{530 GPU-hours} across locally-run models spanning audio-visual, vision-only, and audio-only baselines.

\section{Broader Impacts}
\label{app:broader_impacts}

MMOU is a benchmark designed to evaluate omni-modal understanding in multimodal large language models, and as such carries both positive and negative societal implications. On the positive side, MMOU enables rigorous, structured evaluation of joint audio-visual reasoning — a capability central to real-world AI applications such as accessibility tools (e.g., automated audio description for the visually impaired), educational video analysis, and human-computer interaction. By exposing systematic failure modes in current models, MMOU can guide the development of more robust and equitable multimodal systems. On the negative side, advances in omni-modal understanding enabled by benchmarks like MMOU could accelerate development of systems capable of automated video surveillance, large-scale media monitoring, or generation of convincing multimodal disinformation. We note that our dataset is released solely for non-commercial research purposes.

\section{Dataset Quality Control and Safeguards}
\label{app:qc}

To ensure benchmark quality and content safety, we implemented a multi-stage quality control process. After the initial annotation round, a separate group of reviewers audited 20\% of QA pairs across four axes: (1) relevance of the question to the video, (2) grammatical correctness, (3) accuracy of the assigned skill/task type, and (4) correctness of the provided answer. Questions failing any criterion were revised or discarded, with feedback relayed iteratively between reviewers and annotators until all retained questions met the quality bar.

For content safety, all collected videos were manually reviewed by annotators prior to question generation. Videos containing harmful, offensive, or sensitive content were excluded from the dataset. Videos were sourced from publicly available platforms (e.g., YouTube) and are used in accordance with their Terms of Service for non-commercial research purposes.

\section{Baseline Model Licenses}
\label{app:licenses}

Table~\ref{tab:licenses} lists all baseline models evaluated in this paper along with their associated licenses and references. Closed-source models and API-only LLMs were accessed under their respective terms of service.

\begin{table}[h]
\centering
\caption{Licenses for all baseline models used in MMOU evaluations. AV = Audio-Visual, V = Vision-only, A = Audio-only.}
\label{tab:licenses}
\resizebox{\columnwidth}{!}{%
\begin{tabular}{llll}
\toprule
\textbf{Model} & \textbf{Type} & \textbf{License} & \textbf{Reference} \\
\midrule
\multicolumn{4}{l}{\textit{Closed-Source Audio-Visual MLLMs}} \\
Gemini 2.5 Pro              & AV   & Proprietary (API Terms of Service)          & \citet{comanici2025gemini} \\
Gemini 2.5 Flash            & AV   & Proprietary (API Terms of Service)          & \citet{comanici2025gemini} \\
\midrule
\multicolumn{4}{l}{\textit{Open-Source Audio-Visual MLLMs}} \\
Qwen2.5-Omni-7B             & AV   & Apache 2.0                                  & \citet{xu2025qwen25omni} \\
Qwen3-Omni-30B-Instruct     & AV   & Apache 2.0                                  & \citet{xu2025qwen3omnitechnicalreport} \\
Qwen3-Omni-30B-Thinking     & AV   & Apache 2.0                                  & \citet{xu2025qwen3omnitechnicalreport} \\
Phi-4 Multimodal            & AV   & MIT License                                 & \citet{abouelenin2025phi} \\
Gemma 3n                    & AV   & Gemma Terms of Use                          & \citet{gemmateam2025gemma3technicalreport} \\
MiniCPM-o 4.5               & AV   & Apache 2.0                                  & \citet{openbmb2025minicpm-o2.6} \\
Video-LLaMA 2               & AV   & Apache 2.0                                  & \citet{cheng2024videollama2} \\
OmniVinci                   & AV   & NVIDIA OneWay Noncommercial License         & \citet{ye2025omnivincienhancingarchitecturedata} \\
Baichuan-Omni-1.5           & AV   & Apache 2.0                                  & \citet{li2025baichuan} \\
\midrule
\multicolumn{4}{l}{\textit{Vision-Only MLLMs}} \\
Qwen3-VL-32B-Instruct       & V    & Apache 2.0                                  & \citet{bai2025qwen3vltechnicalreport} \\
Qwen3-VL-8B-Instruct        & V    & Apache 2.0                                  & \citet{bai2025qwen3vltechnicalreport} \\
Qwen2.5-VL-7B-Instruct      & V    & Apache 2.0                                  & \citet{bai2025qwen25vl} \\
\midrule
\multicolumn{4}{l}{\textit{Audio-Only MLLMs}} \\
Audio Flamingo 3            & A    & NVIDIA OneWay Noncommercial License         & \citet{goel2025audio} \\
Qwen3-Omni-30B (audio-only) & A    & Apache 2.0                                  & \citet{xu2025qwen3omnitechnicalreport} \\
\midrule
\multicolumn{4}{l}{\textit{Text-Only LLMs \& Cascaded Models}} \\
Qwen3-235B                  & Text & Apache 2.0                                  & \citet{yang2025qwen3} \\
GPT-4o mini                 & Text & Proprietary (API Terms of Service)          & \citet{openai2024gpt4o} \\
GPT-4.1                     & Text & Proprietary (API Terms of Service)          & \citet{openai2025gpt52systemcard} \\
\bottomrule
\end{tabular}%
}
\end{table}




\end{document}